\documentclass[lettersize,journal]{IEEEtran}
\usepackage{amsmath,amsfonts}
\usepackage{algorithmic}
\usepackage{algorithm}
\usepackage{array}
\usepackage[caption=false,font=normalsize,labelfont=sf,textfont=sf]{subfig}
\usepackage{textcomp}
\usepackage{stfloats}
\usepackage{url}
\usepackage{verbatim}
\usepackage{graphicx}
\usepackage{cite}
\usepackage{xcolor}         % colors
\usepackage{makecell}
\usepackage{colortbl}
\usepackage{pifont}
\usepackage[utf8]{inputenc} % allow utf-8 input
\usepackage[T1]{fontenc}    % use 8-bit T1 fonts
\usepackage{multirow}
\usepackage{tabularx}
\usepackage{array}
\usepackage{amsmath}
\usepackage{booktabs} 
\usepackage{svg}

\begin{document}

\title{How Effective Are Time-Series Models for Precipitation Nowcasting?\\
A Comprehensive Benchmark for GNSS-based Precipitation Nowcasting}

\author{
Yifang~Zhang,
Shengwu~Xiong,
Henan~Wang,
Wenjie~Yin,
Jiawang~Peng,
Yuqiang~Zhang,
Chen~Zhou,
Hua~Chen,
Qile~Zhao,
and~Pengfei~Duan% <-this % stops a space
\thanks{Yifang~Zhang is with the Sanya Science and Education Innovation Park, Wuhan University of Technology, Sanya, 572000, China and also with the School of Computer Science and Artificial Intelligence, Wuhan University of Technology, Wuhan 430070, China (e-mail: yifangzhang@whut.edu.cn).}%
\thanks{Pengfei~Duan, Henan~Wang, and Jiawang~Peng are with the School of Computer Science and Artificial Intelligence, Wuhan University of Technology, Wuhan 430070, China (e-mail: duanpf@whut.edu.cn; 361332@whut.edu.cn; 297975@whut.edu.cn).}%
\thanks{Shengwu~Xiong is with the Interdisciplinary Artificial Intelligence Research Institute, Wuhan College, Wuhan 430212, China, and also with the Shanghai Artificial Intelligence Laboratory, Shanghai 200232, China (e-mail: xiongsw@whut.edu.cn).}%
\thanks{Wenjie~Yin, Yuqiang~Zhang, and Chen~Zhou are with the School of Earth and Space Science and Technology, Wuhan University, Wuhan 430072, China (e-mail: windsoryin@whu.edu.cn; chenzhou@whu.edu.cn; yqzhang\_3@whu.edu.cn).}%
\thanks{Hua~Chen is with the School of Water Resources and Hydropower Engineering, Wuhan University, Wuhan 430062, China (e-mail: chua@whu.edu.cn).}%
\thanks{Qile~Zhao is with the GNSS Research Center, Wuhan University, Wuhan 430062, China (e-mail: zhaoql@whu.edu.cn).}%
\thanks{Corresponding authors: Pengfei Duan (duanpf@whut.edu.cn).}

}

\markboth{IEEE Transactions on Geoscience and Remote Sensing,~Vol.~14, No.~8, October~2025}%
{Zhang \MakeLowercase{\textit{et al.}}: How Effective Are Time-Series Models for Precipitation Nowcasting?}

\maketitle
\begin{abstract}
Precipitation Nowcasting, which aims to predict precipitation within the next 0 to 6 hours, is critical for disaster mitigation and real-time response planning. However, most time series forecasting benchmarks in meteorology are evaluated on variables with strong periodicity, such as temperature and humidity, which fail to reflect model capabilities in more complex and  practically meteorology scenarios like precipitation nowcasting.
To address this gap, we propose \textbf{\textit{RainfallBench}}, a benchmark designed for precipitation nowcasting, a highly challenging and practically relevant task characterized by zero inflation, temporal decay, and non-stationarity, focusing on predicting precipitation within the next 0 to 6 hours. The dataset is derived from five years of meteorological observations, recorded at hourly intervals across six essential variables, and collected from more than 140 Global Navigation Satellite System (GNSS) stations globally. In particular, it incorporates precipitable water vapor (PWV), a crucial
indicator of rainfall that is absent in other datasets.
We further design specialized evaluation protocols to assess model performance on key meteorological challenges, including multi-scale prediction, multi-resolution forecasting, and extreme rainfall events, benchmarking 17 state-of-the-art models across six major architectures on RainfallBench.
Additionally, to address the zero-inflation and temporal decay issues
overlooked by existing models,
we introduce \textbf{\textit{Bi-Focus Precipitation Forecaster (BFPF)}}, a plug-and-play module that incorporates domain-specific priors to enhance rainfall time series forecasting. Statistical analysis and ablation studies validate the comprehensiveness of our dataset as well as the superiority of our methodology. 

% Code and datasets are available at \url{https://anonymous.4open.science/r/RainfallBench-A710}. 
\end{abstract}

\begin{figure*}[!t]
\centering
\includegraphics[width=0.8\textwidth]{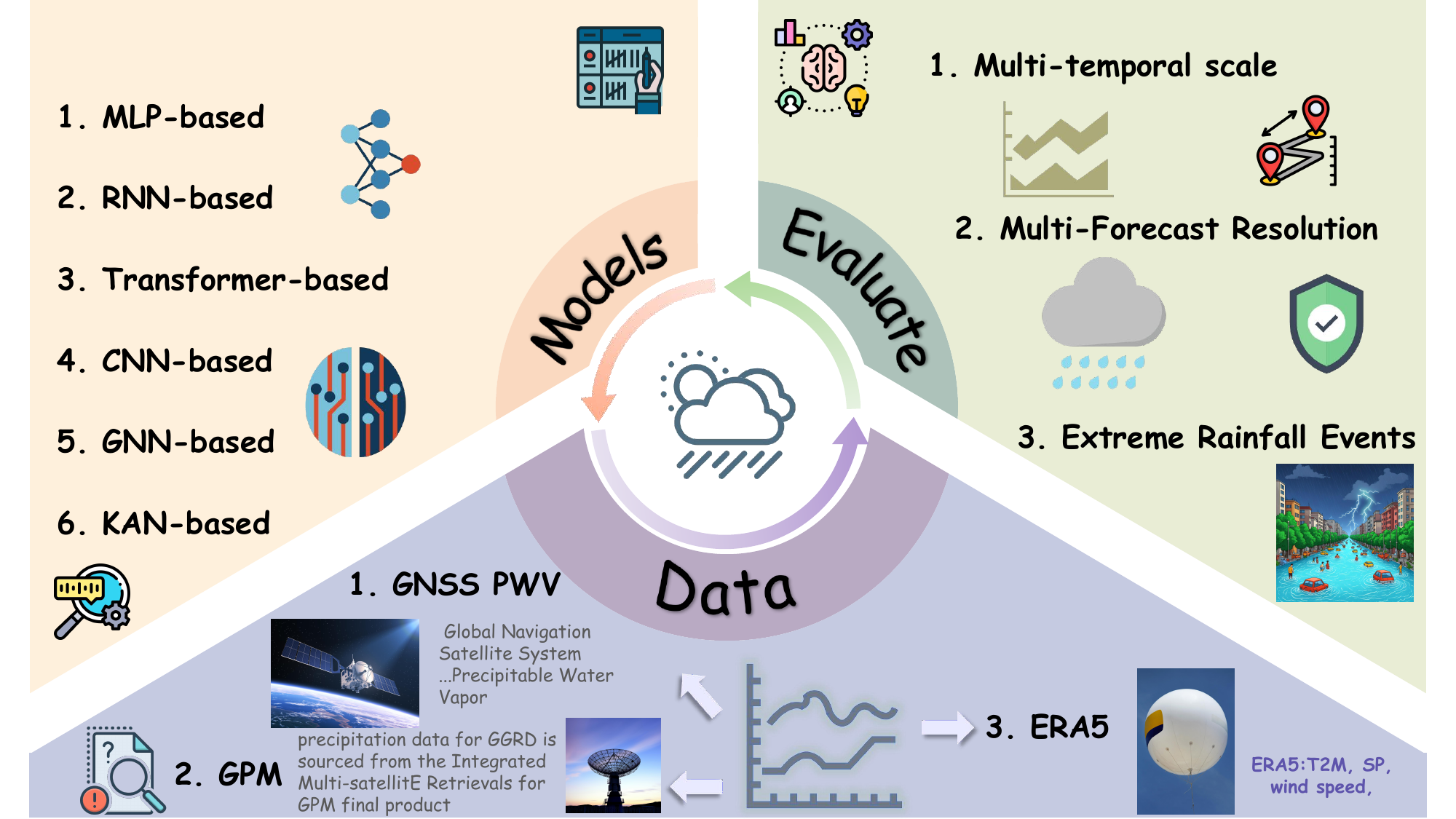}
\caption{Overview of the RainfallBench framework. The benchmark is organized into three main components: the data layer, the model layer, and the evaluation layer. The data layer integrates three sources: GNSS-PWV, ERA5, and GPM. The model layer includes 17 models across six major deep learning architectures, while the evaluation layer encompasses multi-scale prediction, multi-resolution forecasting, and extreme rainfall assessment.} 
\label{fig_intro}
\end{figure*}

\section{Introduction}
Precipitation nowcasting, which focuses on predicting precipitation within the next 0 to 6 hours \cite{zhang2023skilful,nearing2024global}, plays a crucial role in disaster mitigation, flood prevention, and real-time decision-making in weather-sensitive sectors. However, current time series forecasting models in meteorology are often evaluated on variables that exhibit strong periodic patterns, such as temperature and humidity. While these benchmarks facilitate model development and comparison, they often fall short in capturing the complexity and uncertainty inherent in real-world meteorology scenarios, and do not adequately assess model performance on rainfall prediction—one of the most critical atmospheric variables. This gap raises concerns about the practical applicability and robustness of existing models. 

To bridge this discrepancy, we introduce \textbf{RainfallBench}, a benchmark tailored for precipitation nowcasting — a task characterized by zero inflation, temporal decay, and non-stationarity arising from complex atmospheric dynamics. These properties pose substantial challenges to time-series models, making precipitation nowcasting a more realistic and demanding benchmark for evaluating their effectiveness in practical scenarios.

In recent years, rainfall forecasting has spurred intense activity from the deep learning community. On one hand, most precipitation nowcasting methods rely on weather radar imagery \cite{franch2020taasrad19,tang2023postrainbench,rivero2015short,an2025deep}, which is effective but constrained by high costs, limited coverage, and inconsistent continuity. On the other hand, existing benchmarks for time-series forecasting models in meteorology mainly target longer-term multivariate climate variable forecasting \cite{wu2021autoformer,mouatadid2023subseasonalclimateusa} and do not address precipitation nowcasting needs. 

In particular, when using the commonly adopted Weather dataset  \footnote{\url{https://www.bgc-jena.mpg.de/wetter/}} for time series forecasting, most models adopt a multivariate prediction setting, and even in univariate settings, the target variable is typically the last column—CO$_2$ concentration of ambient air. As a result, the evaluation metrics derived from this dataset do not adequately reflect the capability of time series models in the context of rainfall forecasting. Moreover, effective precipitation nowcasting depends on variables strongly correlated with precipitation (e.g., PWV) over the nowcasting horizon, which are often missing from other datasets, making model development and evaluation challenging.
To the best of our knowledge, \textbf{RainfallBench} is the first benchmark dedicated to precipitation nowcasting solely based on historical numerical meteorological records, explicitly incorporating PWV.

It consists of data collected from over 140 GNSS stations globally between 2018 and 2024, covering six key meteorological variables.
All measurements are sampled at hourly intervals, enabling fine-grained temporal modeling.

Specifically, RainfallBench offers several distinctive characteristics that set it apart from existing rainfall forecasting datasets:
\textbf{i) Integration of GNSS-Derived Atmospheric Water Vapor}:
The dataset includes precipitable water vapor (PWV) derived from GNSS observations, which reflects atmospheric moisture and correlates strongly with precipitation onset, making it a key and timely indicator for precipitation nowcasting \cite{10737421}.
\textbf{ii) High-Resolution Temporal Sampling}:
All variables are recorded at hourly intervals, enabling the capture of rapid atmospheric dynamics. This high-frequency sampling improves the suitability of the dataset for precipitation nowcasting within a 0 to 6 hours horizon.
\textbf{iv) Derived from Latest Real-World Scenarios}: Our collected dataset comprises records from 2018-2024, obtained from professional meteorological observation stations, ensuring its strong practical utility.

To ensure a professional evaluation, we propose a more holistic evaluation framework, which assesses models across three key dimensions: \textbf{i) Multi-Time Scale Prediction Evaluation}: Evaluates a model’s rainfall prediction capability across different combinations of input and output sequence lengths.
\textbf{ii) Multi-Forecast Resolution Evaluation}: 
It measures a model's ability to predict rainfall at various temporal granularities. \textbf{iii) Extreme Rainfall Event Evaluation}: It focuses on a model's performance in forecasting sudden, high-intensity rainfall events. 

% \textbf{iii) Accuracy and Reliability of Time Series Prediction}: It comprehensively considers both the precision and stability of the model's predictions. Based on this framework, we benchmark over 20 state-of-the-art deep learning models spanning various architectural families. 

Through a comprehensive evaluation, we identify that existing models often overlook the zero-inflation and temporal decay characteristics, compared to the widely acknowledged non-stationarity of time series data \cite{liu2022non, liu2023koopa, liutimestacker, liutimebridge}. To address these limitations, we design the \textbf{BFPF} to reinforce its sensitivity to rainfall patterns and recent temporal information. Experimental results validate the effectiveness of our approach, offering a new perspective for adapting time series models to precipitation nowcasting.

In summary, our key contributions are as follows:
\begin{itemize}
		\setlength{\itemsep}{0.03em} 
        \item \textbf{Professional Dataset for Precipitation Nowcasting:} 
        Our benchmark is constructed from data collected at hourly intervals between 2018 and 2024 from over 140 GNSS stations globally. It covers six key variables, including PWV, and is specifically curated to support precipitation nowcasting. The dataset will be continuously updated.
        
        \item \textbf{Rainfall-Centric Evaluation Strategy:} We design a tailored evaluation strategy from a meteorological perspective, focusing on multi-scale forecasting, multi-temporal resolution, and extreme rainfall events.

        \item \textbf{Novel plug-and-play  Module for Precipitation Nowcasting:} 
        We introduce the Bi-Focus Precipitation Forecaster, a plug-and-play module that explicitly addresses zero inflation and temporal decay in rainfall data, achieving state-of-the-art performance in extreme rainfall forecasting.
\end{itemize}

\section{Related Works}
\subsection{GNSS-based Precipitation Nowcasting}
In recent years, GNSS-derived Precipitable Water Vapor (PWV) has gained considerable attention for its potential in precipitation nowcasting. Yao et al. \cite{yao2017establishing} proposed a method where a sharp rise in PWV can signal impending rainfall by analyzing hourly data from the Zhejiang Continuously Operating Reference Station (CORS) network between 2014 and 2015. Unlike previous work focused primarily on statistical relationships between PWV and rainfall, Profetto et al. \cite{profetto2025two} Profetto et al. proposed a novel two-step machine learning framework that combines a Random Forest (RF) model with a Long Short-Term Memory (LSTM) neural network, which was validated using data collected between 2021 and 2023 from the GNSS meteorology station located on the roof of the LaMMA Consortium in Sesto Fiorentino, Tuscany. Liu et al. \cite{10942428} proposes a novel deep learning-based model for precipitation nowcasting, which integrates GNSS-derived precipitable water vapor (PWV) data with radar observations.
Lu et al. \cite{11077366} proposed an enhanced precipitation nowcasting model, RSG-GAN, which integrates radar QPE, GOES-16 SWD, and GNSS ZTD data to improve forecasting accuracy over the U.S. west coast.
Yin et al. 
\cite{yin2024lightning} proposed the approach utilized machine learning algorithms to predict lightning occurrences up to 30 minutes in advance.

However, most of these studies are based on a limited number of GNSS stations in local regions, or they focus solely on establishing predictive relationships between PWV as a single variable and rainfall. To enable large-scale validation and fully leverage additional meteorological variables, it is necessary to extend these approaches beyond local datasets and single-variable models.

\subsection{Benchmarks for Time-Series Forecasting}
 Time series models play a crucial role in many fields, and a variety of benchmark datasets and evaluation frameworks have been developed to standardize performance assessment and ensure comparability across studies. 
 
 For instance, FinTSB \cite{hu2025fintsb} emphasizes diversity, standardization, and real-world relevance in financial forecasting. Physiome-ODE \cite{klotergens2025physiome} introduces irregularly sampled ODE-based biological datasets for IMTS evaluation. Cherry-Picking \cite{roque2025cherry} warns against dataset bias and calls for more representative evaluations. TSFM-Bench and GIFT-Eval \cite{aksugift} assess foundation models in zero-, few-, and full-shot regimes. TFB \cite{qiu2024tfb} and TSPP \cite{bkaczek2023tspp} propose unified pipelines to ensure fair and reproducible forecasting. LargeST \cite{liu2023largest} offers a long-term, large-scale traffic dataset with rich metadata to evaluate deep models in realistic settings.

 Despite these advancements, the field of precipitation nowcasting still lacks a comprehensive and standardized benchmarking framework. The absence of such a framework not only hinders fair and reproducible comparisons between models but also limits the systematic evaluation of model generalization under diverse meteorological conditions, which is critical for real-world deployment and operational forecasting.

\subsection{Benchmark for precipitation Nowcasting}
PostRainBench \cite{tang2023postrainbench} introduced a comprehensive multi-variable numerical weather prediction (NWP) post-processing benchmark with a temporal resolution of 3 hours. However, it does not include PWV and is therefore unsuitable for nowcasting applications. RainBench \cite{de2021rainbench} provides a large-scale, multi-modal benchmark using SimSat, ERA5, and IMERG for global precipitation forecasting. Rodriguez Rivero et al. 
Shi et al. \cite{shi2017deep} proposed both a new model and a benchmark for precipitation nowcasting based on radar echo maps from the Hong Kong Observatory.
Ana et al. \cite{an2025deep} provides a comprehensive review of deep learning-based precipitation forecasting methods that utilize multi-source observational data, such as radar reflectivity and satellite imagery.
However, in the domain of precipitation nowcasting based on GNSS-PWV, a comprehensive and systematic benchmark has yet to be established.

The aforementioned related works reveal that current research in GNSS-based precipitation nowcasting is predominantly focused on localized regions, with most studies concentrating on single-factor analysis (e.g., PWV), and a notable lack of research on global-scale, multi-variable integration. Additionally, while significant advancements have been made in time series forecasting across various domains, there remains a scarcity of studies specifically addressing the application of time series models for GNSS-based precipitation nowcasting. Given the increasing use of deep learning models, particularly time series forecasting models, in the field of GNSS precipitation nowcasting, there is a pressing need to construct a globally representative, multi-variable integrated dataset and to establish a robust evaluation framework based on deep learning models to facilitate the further development and application of this area of research.

\section{RainfallBench}
RainfallBench is structured into three main components: the data layer, the model layer, and the evaluation layer. The overall framework is illustrated in Figure \ref{fig_intro}. In the following sections, Section \ref{sec:A} presents a formal problem definition for GNSS-based precipitation nowcasting within the context of time series forecasting. Sections \ref{sec:B}, \ref{sec:C}, and \ref{sec:D} describe the components of the data layer, Section \ref{sec:E} details the model layer, and Section F covers the evaluation layer.

\subsection{Problem Definition}\label{sec:A}
We formulate precipitation nowcasting as a multivariate-to-univariate time series prediction task. Given a sequence of historical observations comprising both meteorological variables and past rainfall values, the goal is to predict future rainfall over a fixed horizon.

Formally, let the input sequence be defined as:
$$
\mathbf{X} = \{ \mathbf{x}_t \}_{t=1}^{T}, \quad \mathbf{x}_t \in \mathbb{R}^D
$$
where $\mathbf{x}_t$ includes meteorological factors (e.g., temperature, humidity, wind) and the rainfall measurement at time $t$, and $D$ denotes the number of input variables and $T$ represents the length of the input sequence

The target is to predict future rainfall values:
$$
\mathbf{y} = \{ y_{T+1}, y_{T+2}, \dots, y_{T+H} \}, \quad y_{t} \in \mathbb{R}
$$
where $H$ is the prediction horizon. Notably, the output is univariate, focusing solely on future rainfall, despite the multivariate nature of the inputs.

This setting captures the practical demands of real-world rainfall forecasting, where complex environmental factors are used to infer a single but highly critical target variable.
\subsection{Data Collection}\label{sec:B}
RainfallBnech integrates high-frequency PWV from the Nevada Geodetic Laboratory (NGL), high-resolution auxiliary meteorological data from ERA5-Land reanalysis, and global precipitation data from the GPM IMERG Final Precipitation product. The detailed information of the three data sources is summarized in the Table \ref{tab:data_source}.

 \begin{table}[!h]
    \centering
    \caption{The details of RainBench}
    \resizebox{\linewidth}{!}{
    \begin{tabular}{cccc}
    \hline 
    Source Product &Key Parameter(s) Used&Spatial Resolution &Temporal Resolution\\
    \hline
    NGL Troposphere Products&	PWV	&Station-wise	&5 minutes\\
    ERA5-Land&Surface Pressure, 2m Temperature, win speed&0.1° x 0.1°&1 hour\\
    GPM IMERG &	Precipitation&0.1° x 0.1°&30 minutes\\
    \hline
    \end{tabular}
    }
    \label{tab:data_source}
\end{table}
\subsubsection{Data Source}

\textbf{GNSS PWV}. The foundational atmospheric measurements for RainfallBnech are sourced from the tropospheric products generated by the Nevada Geodetic Laboratory (NGL) at the University of Nevada. NGL stands as a world leader in the processing of raw GNSS data, providing products for a vast global network that encompasses over 19,000 stations. This unparalleled global coverage enable RainfallBnech to represent a wide array of climatological and geographical regimes.
NGL employs state-of-the-art processing methodologies, utilizing the GipsyX software suite developed at NASA's Jet Propulsion Laboratory and adhering to the latest standards and reference frames from the International GNSS Service (IGS). This ensures the highest possible quality and consistency in the derived products. For the RainfallBnech dataset, we utilize the PWV variable, which are available at high temporal resolutions, including a 5-minute sampling rate for many stations. This high frequency is essential for capturing the rapid temporal evolution of atmospheric water vapor that often precedes precipitation events.\

\textbf{GPM}. The ground-truth precipitation data for RainfallBench is sourced from the Integrated Multi-satellitE Retrievals for GPM (IMERG) final product, specifically Version 07. The Global Precipitation Measurement (GPM) mission is an international satellite constellation designed to provide next-generation observations of rain and snow worldwide. The IMERG Final Run product is selected because it is widely regarded as the highest-quality, research-grade satellite precipitation dataset available. It use the incorporation and calibration of the satellite estimates with data from the Global Precipitation Climatology Centre's (GPCC) network of monthly surface rain gauges. This gauge-correction step significantly reduces biases and improves the overall accuracy of the precipitation estimates, making it the most suitable choice for a benchmark dataset where the quality of the target variable is paramount. The IMERG Final Run provides quasi-global (typically 60°N-S) precipitation estimates at a high spatial resolution of 0.1° x 0.1° and a half-hourly temporal resolution. This fine spatio-temporal sampling is critical for capturing the often localized and short-lived nature of convective rainfall events, which might be missed by coarser products. The specific variable used from the product is precipitation, which provides the calibrated precipitation rate in units of mm/hr. 

\textbf{ERA5-land}. To provide auxilliray meteorological information, surface pressure, temperature and wind speed data are required. For this purpose, RainfallBench incorporates data from ERA5-Land, a global atmospheric reanalysis product generated by the European Centre for Medium-Range Weather Forecasts (ECMWF). ERA5-Land is a replay of the land component of the flagship ERA5 reanalysis, produced using the land surface model. ERA5-Land offers several key advantages that make it the ideal choice for this application. Its primary benefit is its high spatial resolution of approximately 9 km (0.1° x 0.1°), a significant enhancement over the ~31 km grid of the standard ERA5 product. This finer grid is crucial for providing more accurate estimates of surface conditions at the specific locations of the GNSS stations. Furthermore, ERA5-Land provides data at an hourly temporal resolution, which aligns well with the high-frequency nature of the GNSS observations and the need to capture diurnal cycles in atmospheric variables. The dataset also provides a long and consistent historical record, with data available from 1950 to within a few days of the present, enabling the construction of long time series dataset.

\subsubsection{Data Processing}
A critical procedure in creating a multi-source dataset like RainfallBnech is the spatial-temporal alignment of data that exist on different spatial and temporal grids. 

\textbf{Temporal Alignment.} All data streams are aligned to a common hourly temporal grid. The 5-minute NGL ZTD data and 30-minute GPM IMERG data are sampled to produce hourly values centered on the hour, matching the native temporal resolution of the ERA5-Land data.

\textbf{Spatial Alignment.} For the ERA5-Land data, which represent smooth, continuous meteorological fields, a bilinear interpolation method is used. This method estimates the value at the precise latitude and longitude of a given GNSS station by taking a distance-weighted average of the values from the four nearest ERA5-Land grid cells. This approach provides a more accurate local estimate than simply selecting the value of the single nearest grid cell.

In contrast, for the GPM IMERG precipitation data, a nearest neighbor method is employed. The rainfall value assigned to a GNSS station is the value from the GPM grid cell whose centroid is closest to the station's coordinates. Precipitation, especially from convective storms, is a highly discontinuous and highly variational field. Using an interpolation method like the bilinear technique would artificially smooth the data, averaging out high-intensity rainfall cores. This would severely underestimate extreme precipitation events and compromise the dataset's primary utility for nowcasting heavy rain. The nearest neighbor approach, while simpler, better preserves the magnitude and location of rainfall extreme value.

\begin{figure}[htbp]
    \centering
    \includegraphics[width=0.9\linewidth]{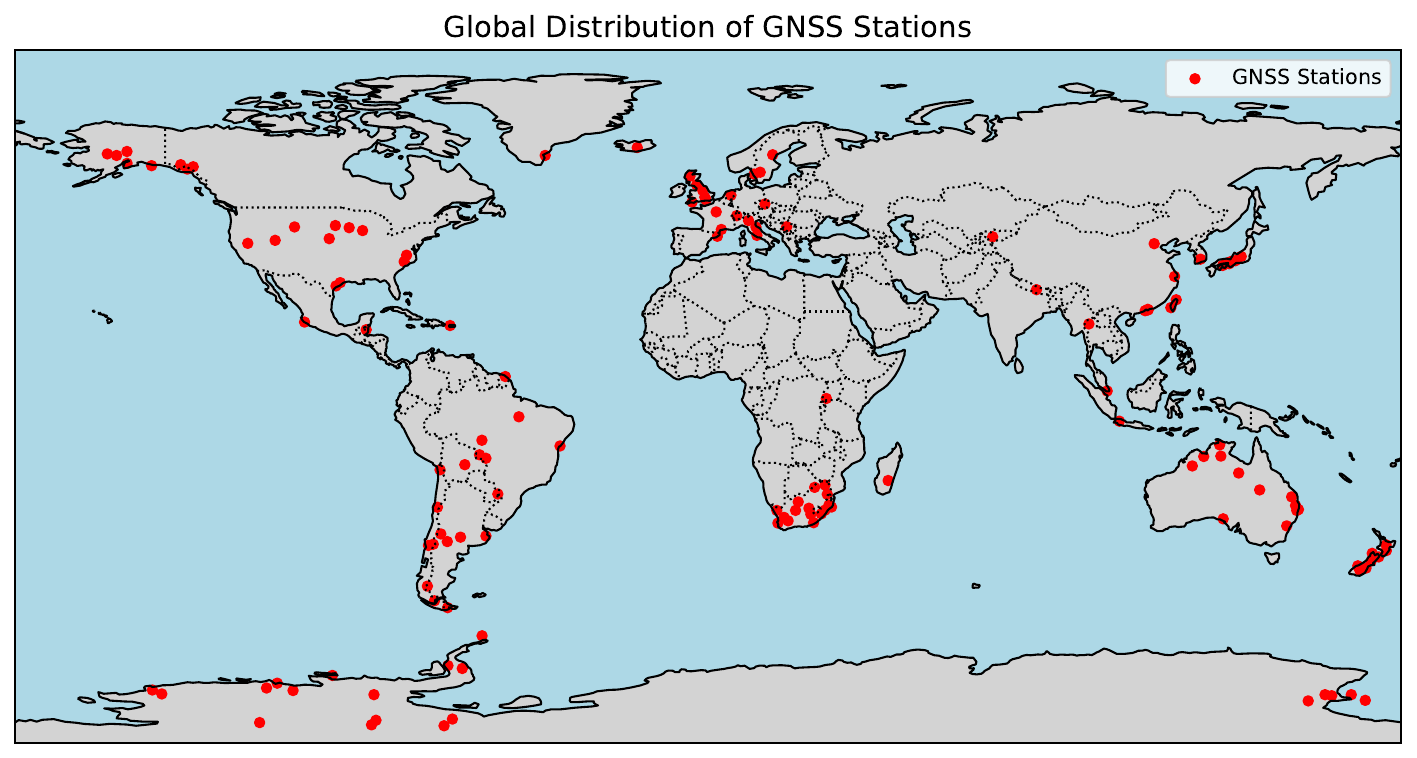}
    \caption{
      Global distribution of 140 selected GNSS stations from the proposed RainfallBench dataset across seven continents, ensuring balanced spatial coverage for evaluating precipitation forecasting models.
    }
    \label{fig:station_distribution}
\end{figure}

\subsection{Quality Control}\label{sec:C}
To ensure data quality, we selected the 20 stations with the highest data completeness from each continent, each of which provides continuous records covering the full period from January 1, 2018, to January 1, 2024, spanning six complete years. For intermediate missing data, we applied a differentiated interpolation strategy to fill in the gaps. For continuous variables such as PWV, T2M, SP, wind speed, and relative humidity, we applied linear interpolation to accurately preserve their spatiotemporal continuity and evolving trends. For precipitation data, forward filling was used to effectively address its discontinuous nature, thus constructing a scientifically sound and robust high-quality dataset. The geographic distribution of the 140 stations is shown in Figure \ref{fig:station_distribution}.

\subsection{Datasets Analysis}\label{sec:D}
        \subsubsection{Data Overview}%%数据集内容的介绍，每一列的含义，及相关性分析
        \begin{figure}[htbp]
    \centering
    \includegraphics[width=0.8\linewidth]{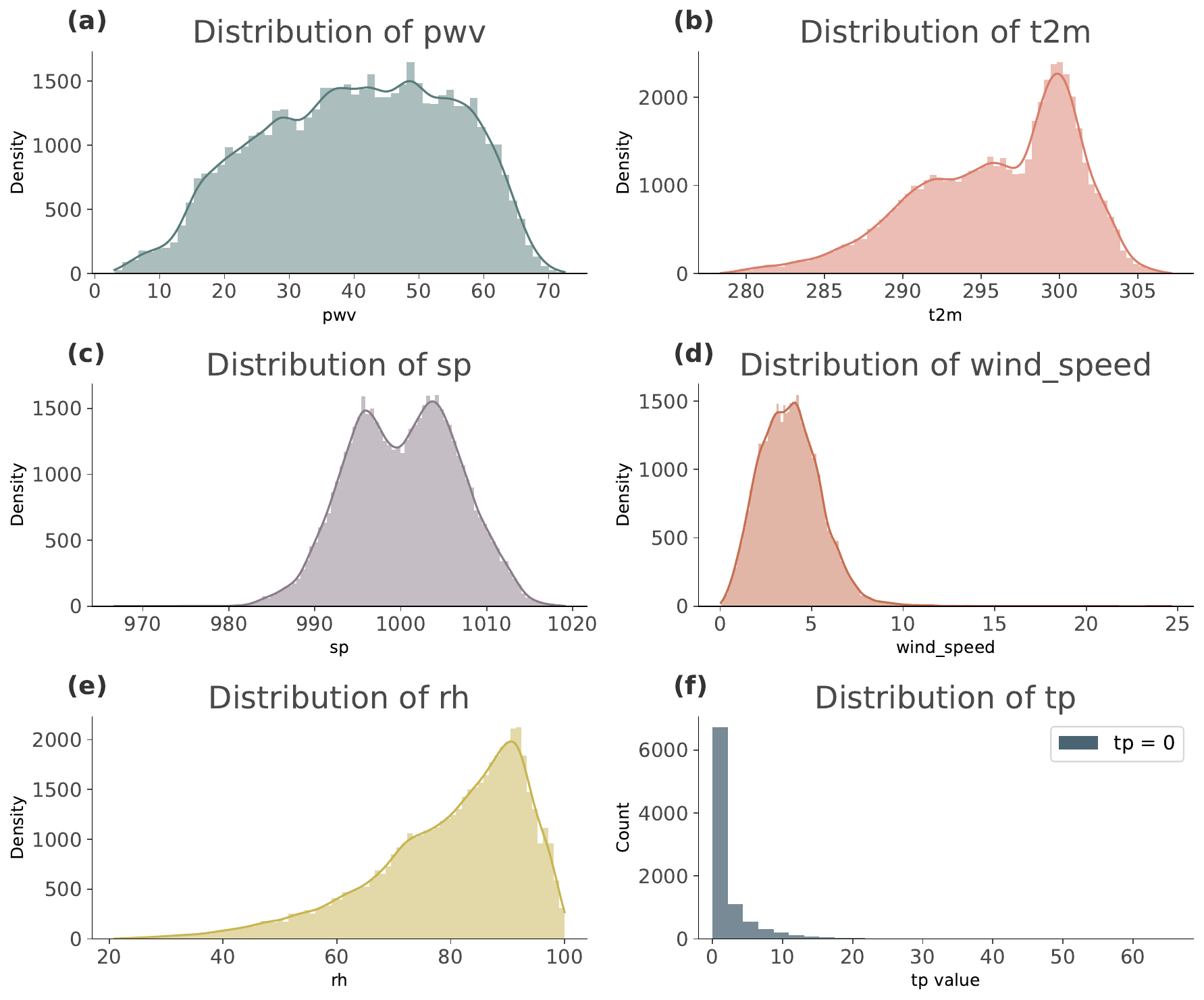}
    \caption{
     Distribution of meteorological variables at the HKST station from 2018 to 2024. The dataset contains six variables per hourly observation.
    }
    \label{fig:variable_distributions}
\end{figure}
Specifically, for the \textit{HKST} station, we utilize a real-world meteorological dataset spanning from January 1, 2018, 00:00 to January 1, 2024, 00:00, with observations recorded every 1 hour, totaling 52,585 time steps without missing entries. Each record consists of six variables (excluding the timestamp): five meteorological features and one target variable representing rainfall. Figure \ref{fig:variable_distributions} illustrates the distribution of values for each variable. Specifically, the input features include:
\begin{itemize}
    \setlength{\itemsep}{0.03em} 
    \item \textbf{t2m}: temperature at 2 meters above ground.
    \item \textbf{sp:} surface pressure
    \item \textbf{rh:} relative humidity
    \item \textbf{wind\_speed:} wind speed.
    \item \textbf{PWV:} precipitable water vapor, retrieved by inverting GNSS signal delays based on their proportional relationship with atmospheric water vapor.
    \item \textbf{tp:} total precipitation (target), obtained from GPM IMERG.
\end{itemize}

\subsubsection{Correlation Analysis}

\begin{figure}[htbp]
    \centering
    \includegraphics[width=0.9\linewidth]{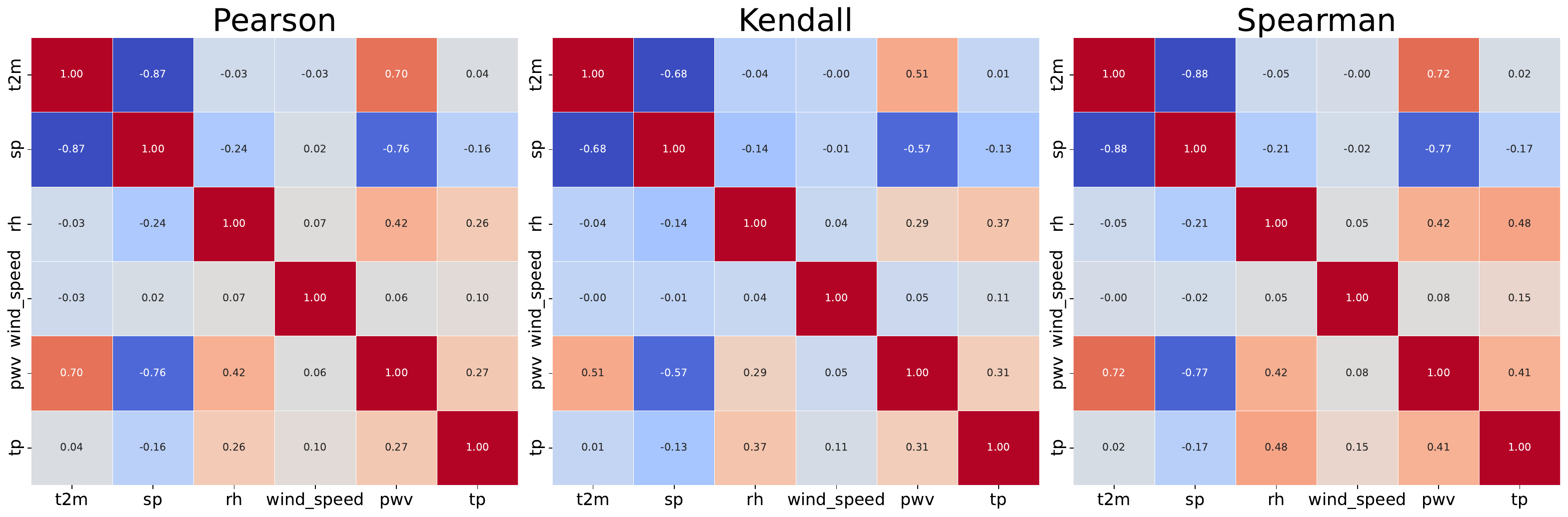}
    \caption{
        Pairwise correlation matrices among meteorological variables and precipitation in the RainfallBench dataset, computed using (a) Pearson, (b) Kendall, and (c) Spearman coefficients. 
    }
    \label{fig:correlation_matrices}
\end{figure}

To explore inter-variable dependencies in the RainfallBench dataset, we perform a correlation analysis using three standard metrics: Pearson, Kendall, and Spearman coefficients. The resulting matrices (Figure~\ref{fig:correlation_matrices}) reveal both linear and monotonic relationships.

Across all correlation matrices, PWV shows the strongest positive correlation with tp (e.g., a Pearson  coefficient of 0.27), highlighting its value as an informative feature for short-term rainfall prediction. In contrast, variables like t2m and sp exhibit weak or negative correlations, indicating limited relevance at nowcasting timescales. Since precipitation nowcasting focuses on the next 0 to 6 hours, model performance hinges on features that reflect rapid and physically meaningful atmospheric changes. Among them, PWV emerges as the most reliable indicator of imminent rainfall, making its inclusion essential in both data selection and model design. 
Moreover, numerous meteorological studies have observed that once PWV exceeds a certain threshold, the probability of precipitation increases significantly \cite{deng2021lenghu, papee1961chemical}.
         \begin{figure}[htbp]
        \centering
\includegraphics[width=0.95\linewidth]{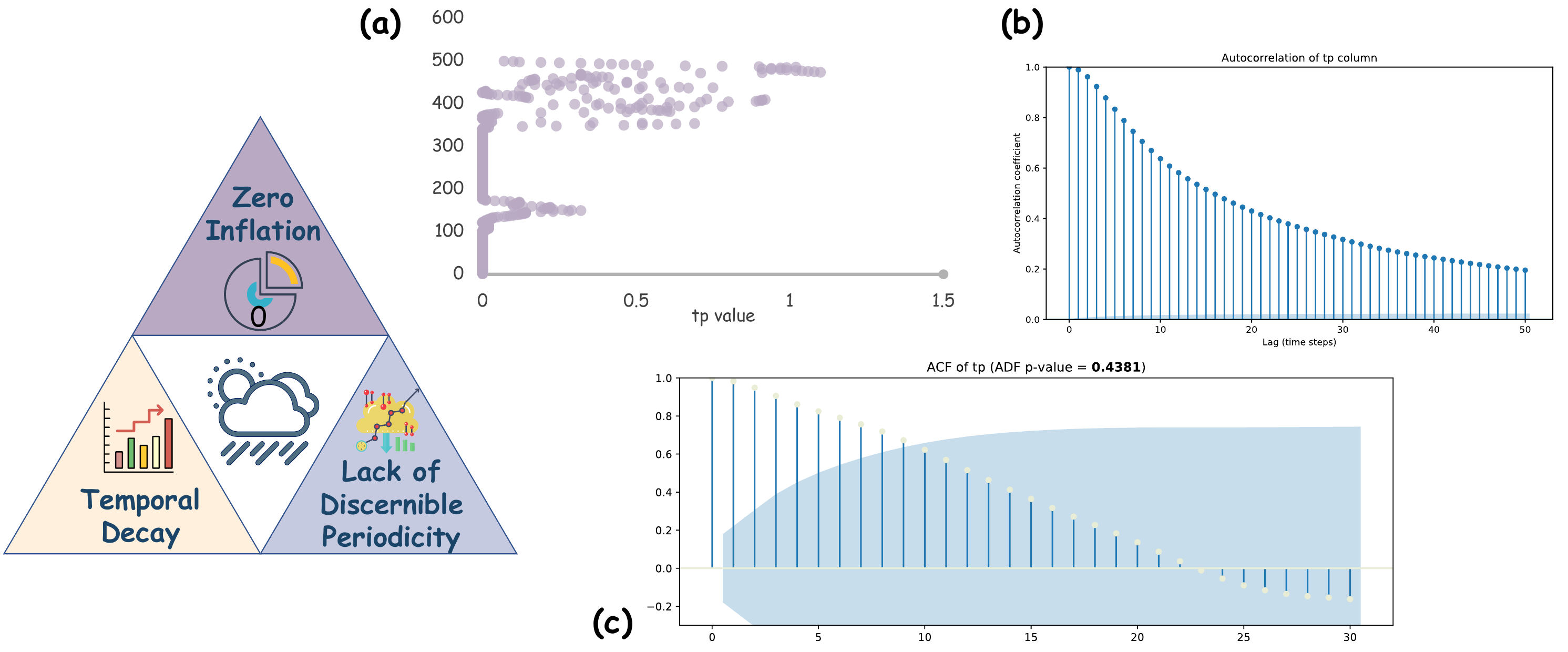}
        \caption{Analysis of data properties in RainfallBench. The benchmark exhibits three key characteristics distinguishing it from standard time series: (i) zero inflation, (ii) temporal dependency decay, and (iii) non-stationarity, along with their implications for modeling.}
        \label{fig:Characteristics}
        \end{figure}

        \subsubsection{Analysis of Data Properties}%%通过图表格式展示其4大特点
        
       RainfallBench introduces three key properties that distinguish it from standard time series benchmarks: (i) zero inflation, (ii) temporal dependency decay, and (iii) non-stationarity. We now analyze each property and its modeling implications.
        
        % 这个图和4个特点的统计图合成一个大图,零膨胀（点分布图），周期性（周期性分析），时间衰减（自相关性分析），真实（位置坐标和accrain图片）
        \textbf{Zero Inflation:} The majority of target values are zeros, reflecting the sparse and event-driven nature of rainfall. This sparsity undermines conventional modeling assumptions that rely on frequent signal continuity. Figure \ref{fig:Characteristics}(a) shows the distribution of 500 randomly sampled records. It is evident that the majority have a tp value of 0. In total, there are 43,313 such records, accounting for 82.3\% of the entire dataset.

        \textbf{Temporal Decay}: When using historical rainfall time series to predict future values, the contributions of past observations vary over time. Typically, more recent data have a stronger influence on the prediction, while the relevance decreases as the time gap widens—an effect we refer to as temporal decay. 

We characterize temporal decay via the lag-$k$ autocorrelation function $\rho(k)$, which quantifies the dependence between current and past rainfall values. Empirically, $\rho(k)$ exhibits an approximately exponential decay:

        \begin{equation}
        \begin{aligned}
\rho(k) \approx e^{-\lambda k}, \quad \lambda > 0
\end{aligned}
\end{equation}
indicating that recent observations carry more predictive information. 
% This pattern aligns with a Markovian assumption, reflecting finite memory in short-term rainfall dynamics.

        This characteristic aligns with the physical nature of rainfall, which tends to evolve gradually rather than starting or stopping abruptly. As shown in the autocorrelation analysis in Figure \ref{fig:Characteristics}(b), this temporal decay pattern is clearly observable.

        \textbf{Non-Stationarity}:
A time series $\{x_t\}_{t=1}^T$ is stationary if its mean, variance, and autocovariance remain constant over time. However, rainfall sequences exhibit strong non-stationarity, particularly in nowcasting contexts, due to fast-changing weather dynamics that lead to rapid shifts in their statistical characteristics. To verify this, we apply the Augmented Dickey-Fuller (ADF) test on a randomly selected segment of 120 values. The test is based on the regression:

\begin{equation}
\begin{aligned}
\Delta x_t = \alpha + \beta t + \gamma x_{t-1} + \sum_{i=1}^{p} \delta_i \Delta x_{t-i} + \varepsilon_t
\end{aligned}
\end{equation}
where $\Delta x_t = x_t - x_{t-1}$ is the first-order difference, $\gamma$ measures the strength of the unit root, and $\varepsilon_t$ is white noise. ADF tests the null hypothesis $H_0 \!: \gamma = 0$ indicates non-stationarity (unit root exists). The resulting p-value of 0.4381 (Figure \ref{fig:Characteristics}(c)) is far above the standard 0.05 significance threshold, failing to reject $H_0$. This confirms the non-stationary nature of the rainfall sequence.

These properties rarely co-occur in other time series datasets, making RainfallBench a uniquely challenging benchmark. It can expose limitations in existing architectures and call for more specialized, domain-adapted solutions. 
\subsection{Comparison Baselines}\label{sec:E}
To ensure a comprehensive benchmark evaluation, we selected 17 models spanning commonly used architectures, including MLP-based, CNN/TCN-based, RNN-based, GNN-based, KAN-based, and Transformer-based designs. To maintain both relevance and rigor, all selected models are state-of-the-art methods proposed in top-tier AI conferences within the past four years. Details of the selected models are summarized in Table~\ref{tab:sota_method_general}.
\subsection{Evaluation Strategy} \label{sec:E}%多时间分辨率，普通MSE，针对大暴雨的，

\textbf{Multi-Temporal Scale Evaluation.}
We evaluate model performance under multiple temporal configurations, considering both the input history length and the forecasting horizon. Formally, we define the set of input lengths as
$$
\mathcal{L}_{\text{in}} = \{12, 24\}
$$
and the set of output lengths (forecasting horizons) as
$$
\mathcal{L}_{\text{out}} = \{2, 4, 6\}
$$corresponding to 1 to 6 hour forecasts in the nowcasting task (1-hour resolution).
Each model is evaluated on all combinations from the Cartesian product $\mathcal{L}_{\text{in}} \times \mathcal{L}_{\text{out}}$. The forecast sequence length $L_{\text{out}}$ is defined as an element of the output length set:
$
L_{\text{out}} \in \mathcal{L}_{\text{out}}.
$

For each setting, we compute both the Mean Squared Error (MSE) and Mean Absolute Error (MAE) between the predicted rainfall sequence $\hat{y}_{1:L_{\text{out}}}$ and the ground truth sequence $y_{1:L_{\text{out}}}$, defined as:

\begin{equation}
\begin{aligned}
\text{MSE} = \frac{1}{L_{\text{out}}} \sum_{t=1}^{L_{\text{out}}} (\hat{y}_t - y_t)^2 \quad
\end{aligned}
\end{equation}
\begin{equation}
\begin{aligned}
\text{MAE} = \frac{1}{L_{\text{out}}} \sum_{t=1}^{L_{\text{out}}} |\hat{y}_t - y_t|
\end{aligned}
\end{equation}

\textbf{Multi-Forecast Resolution Evaluation.}  
In this evaluation, we keep the input data at a fixed temporal resolution of 1 hour, while assessing model performance under different forecast resolutions. Formally, we define the set of forecast resolutions as
$$
\mathcal{R}_{\text{out}} = \{1\text{h}, 2\text{h}, 3\text{h}\}.
$$
$$
L_{\text{out}} = \frac{H}{\mathcal{R}_{\text{out}}}
$$
Each model is evaluated at each forecast resolution in the set $\mathcal{R}_{\text{out}}$. 

For each forecast resolution, we compute both the MSE and MAE between the predicted rainfall sequence $\hat{y}_{1:L_{\text{out}}}$ and the ground truth sequence $y_{1:L_{\text{out}}}$, defined as:

\begin{equation}
\begin{aligned}
\text{MSE} = \frac{1}{L_{\text{out}}} \sum_{t=1}^{L_{\text{out}}} (\hat{y}_t - y_t)^2,
\end{aligned}
\end{equation}

\begin{equation}
\begin{aligned}
\text{MAE} = \frac{1}{L_{\text{out}}} \sum_{t=1}^{L_{\text{out}}} |\hat{y}_t - y_t|.
\end{aligned}
\end{equation}

This evaluation allows us to analyze the model's robustness across different forecast time granularities and to understand how the choice of output temporal resolution affects rainfall prediction performance.

% \textbf{Accuracy and Reliability Evaluation.}  
% Two separate leaderboards are reported accordingly: one for average MSE-based ranking and the other for average MAE-based ranking across all input-output configurations, providing a comprehensive view of each model’s performance in terms of both accuracy and error scale sensitivity.

\textbf{Extreme Rainfall Evaluation.}
% Accurate extreme rainfall forecasting is vital for disaster mitigation. In RainfallBench, we follow the T/CMSA 0013-2019 standard\footnote{\url{http://www.chinamsa.org/uploads/file/20191106142922_61962.pdf}}, under which extreme rainfall is defined as any hourly period with precipitation exceeding 4 mm.
% We label these intervals and evaluate models using MSE and MAE computed only on them.

% \begin{equation}
% \begin{aligned}
% \text{MSE}_{\text{ext}} &= \frac{1}{|E|} \sum_{t \in E} (\hat{y}_t - y_t)^2
% \end{aligned}
% \end{equation}

% \begin{equation}
% \begin{aligned}
% \text{MAE}_{\text{ext}} &= \frac{1}{|E|} \sum_{t \in E} |\hat{y}_t - y_t|
% \end{aligned}
% \end{equation}
% where $E$ denotes the set of time steps labeled as extreme rainfall.
Accurate extreme rainfall forecasting is vital for disaster mitigation. 
In \textbf{RainfallBench}, we follow the T/CMSA 0013-2019 standard\footnote{\url{http://www.chinamsa.org/uploads/file/20191106142922_61962.pdf}}, 
under which extreme rainfall is defined as any hourly period with precipitation exceeding 4 mm. 
We label these intervals and evaluate model performance using the 
\textbf{Extreme Event Reconstruction Error (EERE)} and its absolute variant (\textbf{AEERE}), 
computed only over extreme rainfall periods:

\begin{equation}
\text{EERE} = \frac{1}{|E|} \sum_{t \in E} (\hat{y}_t - y_t)^2
\end{equation}

\begin{equation}
\text{AEERE} = \frac{1}{|E|} \sum_{t \in E} |\hat{y}_t - y_t|
\end{equation}

where $E$ denotes the set of time steps labeled as extreme rainfall events.
A lower EERE or AEERE indicates better reconstruction fidelity of high-intensity precipitation patterns.

\section{Bi-Focus Precipitation Forecaster}
% 需要增加原理图，两条曲线（1条原始注意力曲线，另一条拟合曲线，变成加权后的注意力曲线。在重点位置标记）
\subsection{Motivation}
Our benchmark shows that while existing time series models often consider non-stationarity, they struggle with precipitation nowcasting due to two overlooked domain-specific challenges: zero inflation and temporal decay.

\begin{figure}[htbp]
\centering
\includegraphics[width=0.9\linewidth]{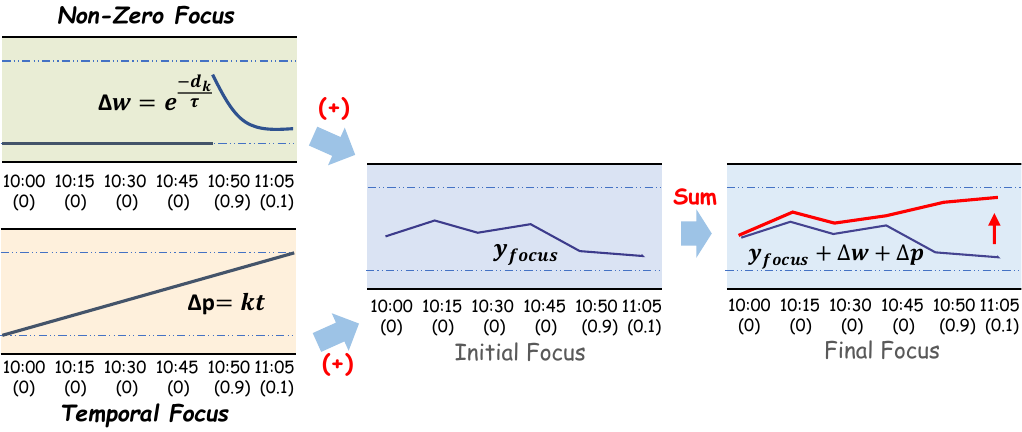}
\caption{Overview of the BFPF module for transformer-based rainfall forecasting. It consists of two key components: (i) Non-Zero Focus, which reduces distractions from non-rainy periods, and (ii) Temporal Focus, which emphasizes temporally proximate context to improve prediction accuracy.}
\label{fig:bifocus}
\end{figure}

To address the challenges in precipitation forecasting, we propose the BFPF, a plug-and-play module for transformer-based models. It consists of two key components: \textbf{Non-Zero Focus} and \textbf{Temporal Focus}, as illustrated in Figure~\ref{fig:bifocus}. The Non-Zero Focus module mitigates distractions caused by non-rainy periods, while the Temporal Focus module emphasizes temporally proximate context.

% Both components are designed based on the intrinsic characteristics of rainfall data and lead to significant improvements in forecasting accuracy.

% Our findings highlight the importance of aligning model design with domain-specific characteristics, moving beyond generic architectures toward more practical and effective forecasting solutions.
\subsection{Non-Zero Focus}
% Non-zero Focus is designed to prevent the model from overlooking
% non-zero values in input sequences dominated by zeros—a
% common issue in rainfall forecasting. Since rainfall events
% are sparse but crucial, treating them equally with zero values
% weakens the model’s focus. Non-zero Focus specifically enhances detection of sudden rainfall spikes within prolonged dry periods, ensuring accurate forecasting of these hydrologically significant events.
The Non-Zero Focus module is designed to mitigate the common challenge in rainfall forecasting where input sequences are dominated by zeros, causing the model to overlook sparse but critical non-zero values. In precipitation data, rainfall events are infrequent yet hydrologically significant. Treating zero and non-zero values equally will dilute the model’s attention to meaningful patterns.

To address this, the Non-Zero Focus enhances the model’s ability to detect sudden rainfall spikes within extended dry periods. It consists of two components: a Non-Zero Context Encoding module that adjusts attention based on value significance, and a Non-Zero Feature Modulation module that reinforces focus on non-zero inputs.

\textbf{Non-Zero Context Encoding}. To guide the attention mechanism toward informative, non-zero values, we introduce a distance-based weighting strategy that quantitatively measures each position’s proximity to the nearest zero. Specifically, for each time step $t$ in the input sequence, we compute the minimal distance to any zero-valued entry:

% \begin{equation}
% \begin{aligned}
% d_t = \min\left(|t - z_l|,\ |z_r - t|\right)
% \end{aligned}
% \end{equation}
\begin{equation}
d_t =
\begin{cases}
+\infty, & \text{if } x_t = 0 \\
\min\left(|t - z_l|,\ |z_r - t|\right), & \text{otherwise}
\end{cases}
\end{equation}
where $z_l$ and $z_r$ denote the indices of the nearest zero positions to the left and right of $t$, respectively. If no zero exists in a given direction, a large sentinel value is used to preserve numerical stability.

The resulting distance matrix $\mathbf{D} \in \mathbb{R}^{B \times L}$ is computed efficiently using a masked cumulative maximum over token positions.

% These distances are then converted into attention bias terms, allowing the model to softly suppress regions close to zero entries while emphasizing non-zero observations.

\textbf{Non-Zero Feature Modulation}. To further refine the model’s sensitivity to informative input regions, we introduce a zero-proximity attention bias that adjusts attention scores based on each key’s distance to the nearest zero. 
% This mechanism acts as a soft prior, discouraging attention toward regions adjacent to uninformative (zero) entries.

Given the previously computed distance matrix $\mathbf{D} \in \mathbb{R}^{B \times L_K}$, we define a proximity weight as:

\begin{equation}
\begin{aligned}
w_k = \exp\left(-\frac{d_k}{\tau}\right)
\end{aligned}
\end{equation}
where $d_k$ is the distance from position $k$ to its nearest zero, and $\tau$ is a temperature hyperparameter controlling decay sharpness. 

These weights are broadcasted and aligned to the attention score tensor $\mathbf{S} \in \mathbb{R}^{B \times H \times L_Q \times L_K}$, and the scores are modulated as:

\begin{equation}
\begin{aligned}
\tilde{\mathbf{S}} = \mathbf{S} + \lambda \cdot \mathbf{W}
\end{aligned}
\end{equation}
where $\lambda$ is a learned scaling factor and $\mathbf{W}$ is the reshaped zero proximity weight matrix. This additive bias encourages the model to assign greater attention to non-zero entries, particularly those representing sudden rainfall onsets, thereby enhancing its focus on rare but meaningful precipitation events.

\subsection{Temporal Focus}
To enhance the attention mechanism with positional awareness, we introduce a linearly increasing positional bias to the original attention scores $\mathbf{S} \in \mathbb{R}^{B \times H \times L_Q \times L_K}$, where $L_K$ is the length of the key sequence. The positional bias vector $\mathbf{p} \in \mathbb{R}^{L_K}$ is defined as:

\begin{equation}
\begin{aligned}
\mathbf{p} = \alpha \cdot \left[ \frac{0}{L_K}, \frac{1}{L_K}, \ldots, \frac{L_K - 1}{L_K} \right]
\end{aligned}
\end{equation}
where $\alpha$ is a learnable scaling factor. This bias is broadcasted to match the shape of $\mathbf{S}$ and added to the attention scores element-wise:

\begin{equation}
\begin{aligned}
\tilde{\mathbf{S}}_{b,h,i,j} = \mathbf{S}_{b,h,i,j} + \mathbf{p}_j
\end{aligned}
\end{equation}

where $b, h, i, j$ index the batch, head, query position, and key position respectively. By explicitly injecting positional information, the model improves its ability to capture the relative ordering of keys, thereby enhancing positional sensitivity in attention computation.
\begin{table*}[t]\setlength\tabcolsep{3pt}
\centering
\caption{Comparison of state-of-the-art methods. The \colorbox{red!50}{red} indicates the best-performing model, while the \colorbox{red!25}{pink} highlights the second-best. Results are obtained with an input sequence length of 24 and an output sequence length of 6.}
\resizebox{0.95\textwidth}{!}{
\begin{tabular}{cc cc cc cc cc cc cc cc cc}
\toprule
\multirow{2}{*}{Methods} & \multirow{2}{*}{\makecell{}{Publication}} & \multicolumn{2}{c}{J340} & \multicolumn{2}{c}{ZIMM} & \multicolumn{2}{c}{P095}& \multicolumn{2}{c}{MTLA}& \multicolumn{2}{c}{ARTA}&
\multicolumn{2}{c}{BFTA} & \multicolumn{2}{c}{FLM5} & \multicolumn{2}{c}{Average} \\ 
\cmidrule(r){3-4} \cmidrule(r){5-6} \cmidrule(r){7-8} \cmidrule(r){9-10} \cmidrule(r){11-12}
\cmidrule(r){13-14} \cmidrule(r){15-16} \cmidrule(r){17-18} 
& & MSE & MAE & MSE & MAE& MSE & MAE& MSE & MAE& MSE & MAE& MSE & MAE & MSE & MAE & MSE & MAE\\
\midrule
\multicolumn{10}{l}{\textbf{MLP-based}} \\ 
\midrule
\multirow{1}{*}{DLinear} & \multirow{1}{*}{AAAI 2023} &1.3654 & 0.4574 & 2.0124& 0.4471 &0.3804&0.1159&1.6673&0.4204&4.0069&0.5785&0.8255&0.2680&0.0000&0.0000 & 1.4654 & 0.3250 \\   
\multirow{1}{*}{Koopa} & \multirow{1}{*}{NIPS 2023} &1.4559 & 0.3525 &2.0211& 0.3716& 0.3896 &0.1030 & 1.7539& 0.3271&4.0590&0.4896&0.8442&0.2272&0.0000&0.0000 &1.5034&0.2673 \\   
\multirow{1}{*}{FilterTS} & \multirow{1}{*}{AAAI 2025} &1.4796 & 0.3573&2.0549&0.3656 &0.3905& 0.1064&1.7334&0.3224&4.1174&0.4916&0.8438&0.2242&0.0000&0.0000&1.5171&0.2668\\   
\midrule
\multicolumn{10}{l}{\textbf{RNN-based}} \\ 
\midrule
\multirow{1}{*}{SegRNN} & \multirow{1}{*}{Arxiv 2023} &\cellcolor{red!25}1.3102 & 0.4228&2.0623&0.4326& 0.4810& 0.1102&1.6353&0.3850&4.3240&0.5598&0.8413&0.2634&0.0000&0.0000&1.5220&0.3105\\   
\multirow{1}{*}{xLSTM} & \multirow{1}{*}{NIPS 2024} &\cellcolor{red!50}1.3002 &0.4240&\cellcolor{red!50}1.9704&0.3838 & \cellcolor{red!25}0.3625& 0.1144&1.5989&0.4228&\cellcolor{red!25}3.7109&0.5639&\cellcolor{red!25}0.7573&0.2500&0.0000&0.0044 &\cellcolor{red!25}1.3857&0.3090\\ 
\multirow{1}{*}{P-sLSTM} & \multirow{1}{*}{AAAI 2025} &1.3368&0.4073&2.0091&0.4278&0.3693&0.1096&1.6303&0.3777&3.8603&0.5651&0.7979&0.2584&0.0000&0.0000&1.4277&0.3066\\   
\midrule
\multicolumn{10}{l}{\textbf{TCN\&CNN-based}} \\ 
\midrule
\multirow{1}{*}{TimesNet} & \multirow{1}{*}{ICLR 2023} &1.4781 & 0.3538 &2.0045&0.3700& 0.3930 &0.1025&1.7046&0.3336&4.0875&0.4962&0.8470&0.2329&0.0000&0.0000&1.5021&0.2700\\   
\multirow{1}{*}{TimeMixer++}& \multirow{1}{*}{ICLR 2025} &1.4825 & 0.3416&2.0404&0.3560 & 0.3906& 0.0956&1.7141&0.3243&4.0926&0.4754&0.8567&0.2245&0.0000&0.0000&1.5110&0.2596\\ 
\multirow{1}{*}{xPatch}& \multirow{1}{*}{AAAI 2025}&1.4169 & 0.3261&2.0269&\cellcolor{red!50}0.3456& 0.3815&\cellcolor{red!25}0.0924&1.6956&0.3245&4.0443&0.4647&0.8354&0.2143&0.0000&0.0000&1.4858&0.2525\\   
\midrule
\multicolumn{10}{l}{\textbf{GNN-based}} \\ 
\midrule
\multirow{1}{*}{MSGNet}& \multirow{1}{*}{AAAI 2024} &1.4417 &0.3451&\cellcolor{red!25}1.9761&0.3657& 0.3828& 0.1042&1.6934&0.3279&4.0096&0.4854&0.8436&0.2262&0.0000&0.0000&1.4782&0.2650\\   
\multirow{1}{*}{TimeFilter} &\multirow{1}{*}{ICML 2025}&1.4128 & 0.3247&2.0217&0.3510&0.3857& 0.0936&1.6920&\cellcolor{red!50}0.3078&4.0255&\cellcolor{red!25}0.4585&0.8318&\cellcolor{red!25}0.2141&0.0000&0.0000&1.4814&\cellcolor{red!50}0.2499\\
\midrule
\multicolumn{10}{l}{\textbf{KAN-based}} \\  
\midrule
\multirow{1}{*}{TimeKAN} &\multirow{1}{*}{ICLR 2025}&1.4101&\cellcolor{red!50}0.3217&2.0198&\cellcolor{red!25}0.3495& 0.3780&0.0933&1.7112&0.3179&4.0501&0.4641&0.8361&\cellcolor{red!25}0.2141&0.0000&0.0000&1.4865&0.2515\\  
\midrule
\multicolumn{10}{l}{\textbf{Transformer-based}} \\ 
\midrule            
\multirow{1}{*}{Informer}& \multirow{1}{*}{AAAI 2021} &1.3601 & 0.3707& 2.0258&0.5461& 0.3913 &0.1287 &\cellcolor{red!25}1.5952&0.3995&3.8286&0.6883&0.7757&0.2470&0.0000&0.0006&1.4252&0.3401\\  
\multirow{1}{*}{PatchTST}& \multirow{1}{*}{ICLR 2023} 
&1.4298 & \cellcolor{red!25}0.3230& 2.0281& 0.3538& 0.3813 &0.0926 & 1.7120&\cellcolor{red!25}0.3105&4.0303&\cellcolor{red!50}0.4582&0.8363&0.2212&0.0000&0.0000&1.4883&\cellcolor{red!25}0.2513\\ 
\multirow{1}{*}{iTransformer} & \multirow{1}{*}{ICLR 2024} &1.4752& 0.3462 &2.0870& 0.3705& 0.3909 &0.1000 & 1.7288&0.3244&4.0953&0.4811&0.8759&0.2497&0.0000&0.0000&1.5217&0.2674\\
\multirow{1}{*}{TimeXer}& \multirow{1}{*}{NIPS 2024} 
&1.6521 & 0.3806& 2.1916& 0.3848& 0.3860 &0.1025 & 1.8817&0.3563&4.2324&0.5082&0.8942&0.2417&0.0000&0.0000&1.6054&0.2819\\
\multirow{1}{*}{PPDformer}& \multirow{1}{*}{ICASSP 2025} &1.4316 & 0.3390 & 2.0907&0.3666&0.3900 &0.0971&1.7307&0.3216&4.1038&0.4734&0.8451&0.2354&0.0000&0.0000&1.5131&0.2618\\   
\multirow{1}{*}{Informer(with BFPF)}& \multirow{1}{*}{Ours}&1.3671&0.3878&2.2093&0.4783&\cellcolor{red!50}0.0901&\cellcolor{red!50}0.0565&\cellcolor{red!50}1.2203&0.3324&\cellcolor{red!50}3.3670&0.6555&\cellcolor{red!50}0.4860&\cellcolor{red!50}0.1929&0.0000&0.0003&\cellcolor{red!50}1.2485&0.2730\\ 
\midrule  
% Average&\textbackslash&1.4409 & 0.3647 & 2.0364 & 0.3867 & 0.3867 & 0.1056 & 1.6863 & 0.3543 & 4.0235 & 0.5081 & 0.8356 & 0.2349 & 0.0000 & 0.0000&\textbackslash&\textbackslash\\
Average&\textbackslash&1.4226 & 0.3656 & 2.0473 & 0.3926 & 0.3730 & 0.1010 & 1.6722 & 0.3464 & 4.0025 & 0.5199 & 0.8152 & 0.2336 & 0.0000 & 0.0003 & \textbackslash&\textbackslash\\
\bottomrule
\end{tabular}
}
\label{tab:sota_method_general}
\end{table*}
\begin{figure*}[htbp]
    \centering
    \includegraphics[width=0.9\textwidth]{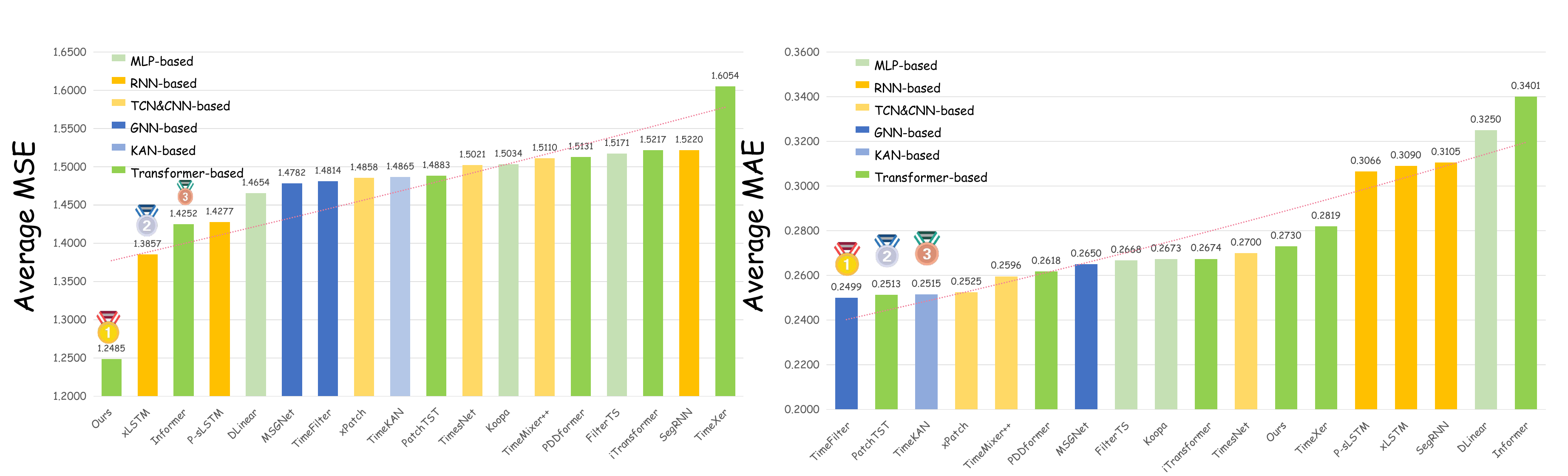}
    \caption{
        Model ranking based on average performance. The left panel shows the ranking by average MSE, while the right panel shows the ranking by average MAE. 
    }
    \label{fig:results}
\end{figure*}
    \section{Experiments}
    \subsection{Experiment Setup}
All experiments are conducted on an NVIDIA H100 80GB GPU. To ensure the generalizability of the experiment, we selected a representative station from each continent for the study. To ensure evaluation consistency, all results are computed using de-normalized actual rainfall values. For robustness, each experiment is repeated three times, and the average performance is reported as the final result. The datasets were split into training, validation, and test sets in a 7:1:2 ratio.

The seven GNSS stations used in experiments are geographically distributed across different continents, elevations, and climate zones. Their details are summarized below:

\textbf{J340 (34.406°N, 135.364°E, 91.983 m)} – Located in the Kinki region of Japan, this station is situated at a low elevation. It is characterized by a humid subtropical climate (Cfa), with four distinct seasons, hot and rainy summers, mild winters, and relatively evenly distributed precipitation throughout the year. The region is also occasionally affected by typhoons in summer. This station has a completeness rate of 99.82\%.

\textbf{ZIMM (46.877°N, 7.465°E, 956.34 m)} – Located in central Switzerland in the Alps, this mid-altitude station exhibits a temperate continental climate (Dfb), characterized by cold, snowy winters and warm, humid summers, with precipitation distributed throughout the year and frequent summer thunderstorms. This station has a completeness rate of 99.85\%.

\textbf{P095 (39.698°N, -119.537°W, 1608.804 m)} – Located in western Nevada, USA, this high-altitude station experiences a temperate desert climate (BWk), with arid conditions, large diurnal temperature variations, cold winters, hot summers, and low annual precipitation. This station has a completeness rate of 99.82\%.

\textbf{MTLA (-15.228°S, -59.35°W, 267.63 m)} – Located in Mato Grosso, Brazil, this low-elevation station exhibits a tropical wet and dry climate (Aw/Am), with a pronounced wet season in summer (November–March) and a dry winter season. The region experiences high average annual temperatures. This station has a completeness rate of 99.64\%.

\textbf{ARTA (-38.618°S, 176.136°E, 369.779 m)} – This station in the eastern North Island of New Zealand lies at a moderate elevation. The climate is temperate oceanic (Cfb), with mild and humid conditions year-round, evenly distributed precipitation, warm summers, and cool winters, strongly influenced by the surrounding ocean. This station has a completeness rate of 99.48\%.

\textbf{BFTA (-29.111°S, 26.205°E, 1441.266 m)} – Situated in northern South Africa, this high-altitude station experiences a subtropical highland climate (Cwb), with warm and wet summers, cool and dry winters, and most precipitation occurring during the summer months. This station has a completeness rate of 98.16\%.

\textbf{FLM5 (-77.533°S, 160.271°E, 1869.726 m)} – Situated on Mount Fleming, Antarctica, this high-elevation polar station is characterized by an ice cap climate (EF). It experiences extremely cold temperatures year-round, very low precipitation (mostly snow), strong winds, and a permanently frozen environment. This station has a completeness rate of 99.77\%.

\subsection{Experiment Results}
This analysis is based on the performance of various time series forecasting models in predicting rainfall at multiple GNSS stations (J340, ZIMM, P095, MTLA, ARTA, BFTA, FLM5). The evaluation metrics include MSE and MAE, where smaller values indicate higher prediction accuracy of the models.
Comprehensive forecasting results are listed in Table \ref{tab:sota_method_general} with the best in \colorbox{red!50}{red} and the second in \colorbox{red!25}{pink}. The lower MSE/MAE indicates the more accurate prediction result. 

Then, our analysis of the experimental results is guided by the following five research questions:

\textbf{RQ1: }How do different model architectures perform in precipitation nowcasting, and which type achieves the best results?(\textbf{\textit{Analysis from the Model Perspective}})

\textbf{RQ2: }How dose various model performance in each area? (\textbf{\textit{Analysis from the Dataset Perspective}})

\textbf{RQ3: }Do our proposed Bi-Focus Precipitation Forecaster improve the performance of Transformer-based models? (\textbf{\textit{Effect Analysis of Bi-Focus Precipitation Forecaster}})

\textbf{RQ4: }How does model performance vary with changes in forecasting horizon and predict length? (\textbf{\textit{Multi-temporal scale Evaluation}})

\textbf{RQ5: }How does model performance vary with changes in multi-forecast resolution? (\textbf{\textit{Multi-Forecast Resolution Evaluation}})

\textbf{RQ6: }How do different models perform for extreme rainfall forecasting?  (\textbf{\textit{Extreme Rainfall Evaluation}})

\subsection{RQ1: Analysis from the Model Perspective}
The MLP-based models, such as DLinear, Koopa, and FilterTS, exhibit moderate performance across most sites, particularly at the J340 and P095 stations, where the MSE and MAE values are relatively high, indicating noticeable prediction errors.
RNN-based models, including SegRNN, xLSTM, and P-slSTM, demonstrate superior performance at several sites, particularly at the J340 and P095 stations. Among these, the xLSTM model achieves the lowest MSE, suggesting its high prediction accuracy at six stations.
TCN and CNN-based models, such as TimesNet, TimeMixer+, and xPatch, perform exceptionally well at the ZIMM and P095 station, with the xPatch model yielding the lowest MAE values, highlighting its superior performance at this site.
GNN-based models, including MSGNet and TimeFilter, show good performance at the MTLA and BFTA stations, with the TimeFilter model achieving the lowest MAE values at the ZIMM station.
The KAN-based model, TimeKAN, demonstrates outstanding performance across multiple sites, especially at the J340, BFTA and P095 stations, where it records the lowest MAE values, indicating the highest prediction accuracy at these sites.
Transformer-based models also exhibit strong performance at multiple sites, with the Informer model achieving the second lowest average MSE among all the models. PatchTST model achieving the second lowest average MAE among all the models.

Based on the average results across all stations, Informer with our proposed BFPF achieves the lowest MSE, indicating its superior overall predictive accuracy, while xLSTM ranks second.
In terms of average MAE, TimeFilter attains the best performance, followed by PatchTST, demonstrating their strong capability in reducing absolute prediction errors.

\subsection{RQ2: Analysis from the Dataset Perspective}
From a dataset perspective, the predictive performance across different sites reflects the distinct characteristics of their rainfall time series. J340 exhibits generally low errors, with MSE ranging from 1.30 to 1.50 and MAE from 0.30 to 0.50, indicating a relatively stable series with few extreme events, high data quality, and low prediction difficulty. ZIMM shows moderately high errors, with MSE exceeding 2.00, primarily due to pronounced seasonality and moderate precipitation events, reflecting a certain level of data complexity. P095 has low MSE values (0.3–0.5), suggesting sparse rainfall and relatively simple prediction conditions. MTLA demonstrates moderate errors, consistent with tropical rainfall concentrated in the wet season and minimal zero-inflation, making predictions relatively manageable. ARTA exhibits high errors, with model MSE generally above 4, indicating a challenging prediction task. BFTA shows relatively low MSE values between 0.75 and 0.90, reflecting localized rainfall patterns and moderate prediction difficulty. Finally, FLM5 presents near-zero errors (MSE $\approx$ 0.0000), as Antarctic precipitation is minimal and the time series is almost entirely zero, rendering the prediction task trivial.

\subsection{RQ3: Effect Analysis of Bi-Focus Precipitation Forecaster}    
\begin{table}[!h]
    \centering
    \caption{Ablation analysis of the Non-Zero \& Temporal Focus modules at the BFTA station. “24(2)” denotes an input sequence length of 24 and an output sequence length of 2; other notations follow the same convention. The \colorbox{red!50}{red} indicates the best-performing model.}
    \resizebox{\linewidth}{!}{
    \begin{tabular}{cc cc cc cc cc cc}
    \hline 
    \multicolumn{2}{c}{Module} & \multicolumn{2}{c}{24(2)} & \multicolumn{2}{c}{24(4)} & \multicolumn{2}{c}{24(6)}\\
    \cline{1-2}
    Non-zero Focus& Temporal Focus&MSE&MAE&MSE&MAE&MSE&MAE\\
    \hline
    \ding{55} & \ding{55} & 0.6343&0.2004 & 0.8047&0.2358 & 0.7757&0.2470 \\
     \ding{55} & \checkmark & \cellcolor{red!50}0.3890&0.1615 & 0.4771&0.1844 & \cellcolor{red!50}0.4801&0.2016 \\
   \checkmark & \ding{55} & 0.3910&0.1655&0.4618&0.1784&0.4800&0.1964\\
    \checkmark & \checkmark &0.3936&\cellcolor{red!50}0.1592 &\cellcolor{red!50}0.4564&\cellcolor{red!50}0.1800&0.4860&\cellcolor{red!50}0.1929 \\
    \hline
    \end{tabular}
    }
    \label{tab_ablation_BFTA}
\end{table}

\begin{table}[!h]
\centering
\caption{Ablation analysis of the Non-Zero \& Temporal Focus modules at the P095 station. “24(2)” denotes an input sequence length of 24 and an output sequence length of 2; other notations follow the same convention. The \colorbox{red!50}{red} indicates the best-performing model.}
\resizebox{\linewidth}{!}{
\begin{tabular}{cc cc cc cc cc cc}
\hline 
\multicolumn{2}{c}{Module} & \multicolumn{2}{c}{24(2)} & \multicolumn{2}{c}{24(4)} & \multicolumn{2}{c}{24(6)}\\
\cline{1-2}
Non-zero Focus& Temporal Focus&MSE&MAE&MSE&MAE&MSE&MAE\\
\hline
\ding{55} & \ding{55} & 0.3449&0.1031 & 0.3677&0.1015 & 0.3913& 0.1287 \\
 \ding{55} & \checkmark &0.0815&0.04971& 0.0867&0.0537&0.0887&0.0522 \\
\checkmark & \ding{55} &0.0826&0.0510 &0.0905&\cellcolor{red!50}0.0537&\cellcolor{red!50}0.0884&0.0571\\
\checkmark & \checkmark & \cellcolor{red!50}0.0798&\cellcolor{red!50}0.0467 & \cellcolor{red!50}0.0857&0.0584 & 0.0901&\cellcolor{red!50}0.0565\\
\hline
\end{tabular}
}
\label{tab_ablation_P095}
\end{table}

     Transformer-based models have gained widespread attention in recent years as powerful tools for sequential modeling across various domains. However, from Table \ref{tab:sota_method_general}, it is evident that Transformer-based models exhibit relatively unstable performance in rainfall time-series forecasting. This limitation arises because standard Transformer architectures fail to explicitly capture key rainfall data characteristics, including high sparsity and rapid temporal decay.
     
     To validate the effectiveness of the proposed BFPF module, we select Informer, one of the strongest Transformer-based models, as our baseline. To further assess the generalization capability of the module, ablation experiments are conducted on two representative stations, P095 and BFTA, under multiple forecasting horizons.
     
     Our ablation study in Table \ref{tab_ablation_BFTA} and Table \ref{tab_ablation_P095} demonstrates that incorporating the proposed BFPF significantly enhances the Transformer’s performance. Incorporating either module individually yields noticeable gains over the baseline, and combining both achieves the best results across multiple forecast horizons. This highlights the importance of customizing attention mechanisms to better model rainfall-specific temporal patterns, thereby enhancing the effectiveness of Transformer models in this task.

\subsection{RQ4: Multi-temporal scale Evaluation}

To ensure a fair and comprehensive evaluation, we selected representative models from six widely-used architectural families in time-series forecasting and conducted experiments on six stations excluding FLM5, as the MSE values at FLM5 are all zero and thus not informative for analysis.
\begin{figure}[htbp]
    \centering
    \includegraphics[width=0.95\linewidth]{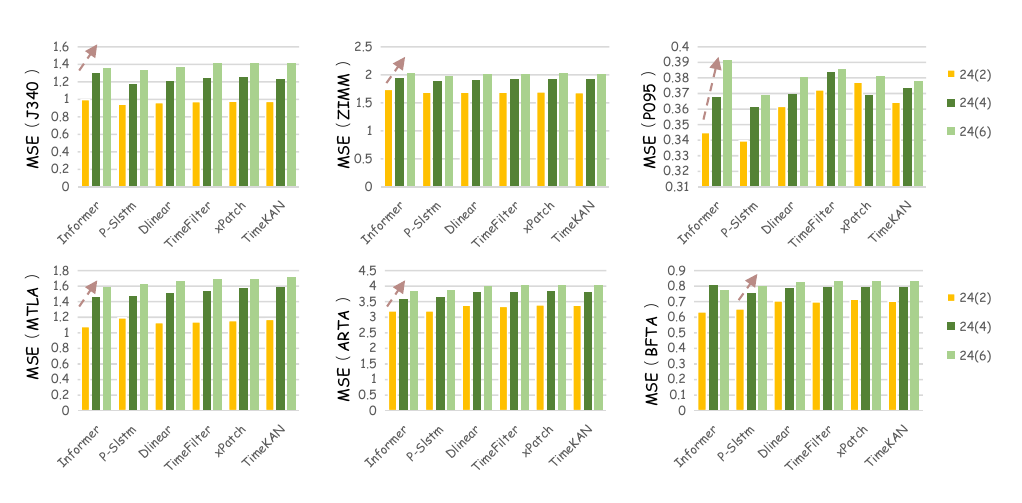}
    \caption{
Model Performance Comparison Across Multiple Temporal Scales (MSE). Fixing the input length at 24 steps and vary the forecast horizon.
    }
    \label{fig:multi-scale-MSE1}
\end{figure}

We consider two evaluation settings. In the first setting, we fix the input length at 24 steps and vary the forecast horizon. As expected, model errors tend to increase with longer prediction horizons. This can be attributed to the compounding uncertainty as the forecast period extends, leading to increased prediction difficulty and higher model error.

\begin{figure}[htbp]
    \centering

    \includegraphics[width=0.95\linewidth]{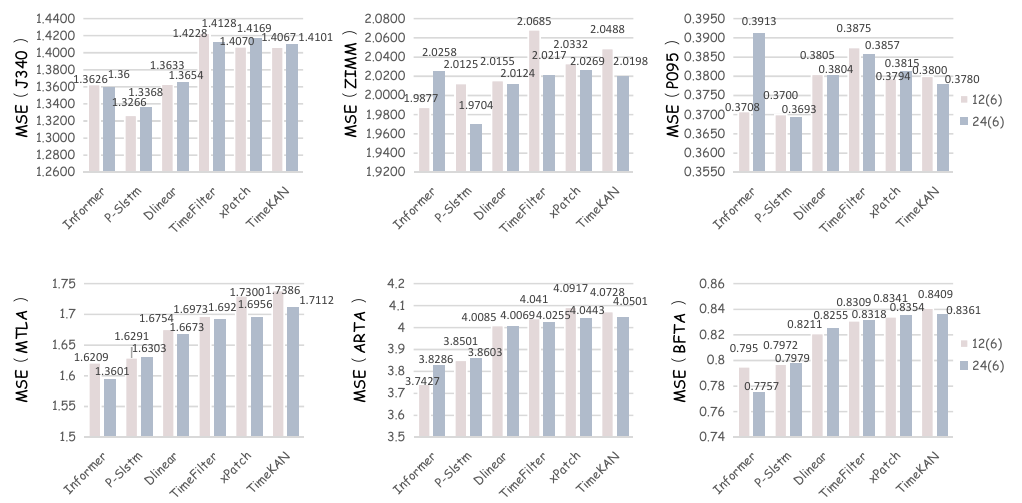}
    \caption{
Model Performance Comparison Across Multiple Temporal Scales (MSE). Fixing the output length at 6 steps and vary the input length (12 vs. 24).
    }
    \label{fig:multi-scale-MSE2}
\end{figure}
In the second setting, we fix the output length at 6 steps and vary the input length (12 vs. 24). We observe that, for most models, increasing the input horizon leads to lower MSE. This suggests that a longer historical context provides more information, which helps improve forecasting accuracy by capturing longer-term trends and dependencies in the time-series data.

However, it is worth noting that for some models, performance actually decreases as the input length increases. This may be due to overfitting to irrelevant or noisy information in the extended input sequence, or the inability of certain architectures to effectively leverage longer temporal dependencies. We will explore this limitation for future work.

\subsection{RQ5: Multi-Forecast Resolution Evaluation}

\begin{figure}[htbp]
    \centering
    \includegraphics[width=0.9\linewidth]{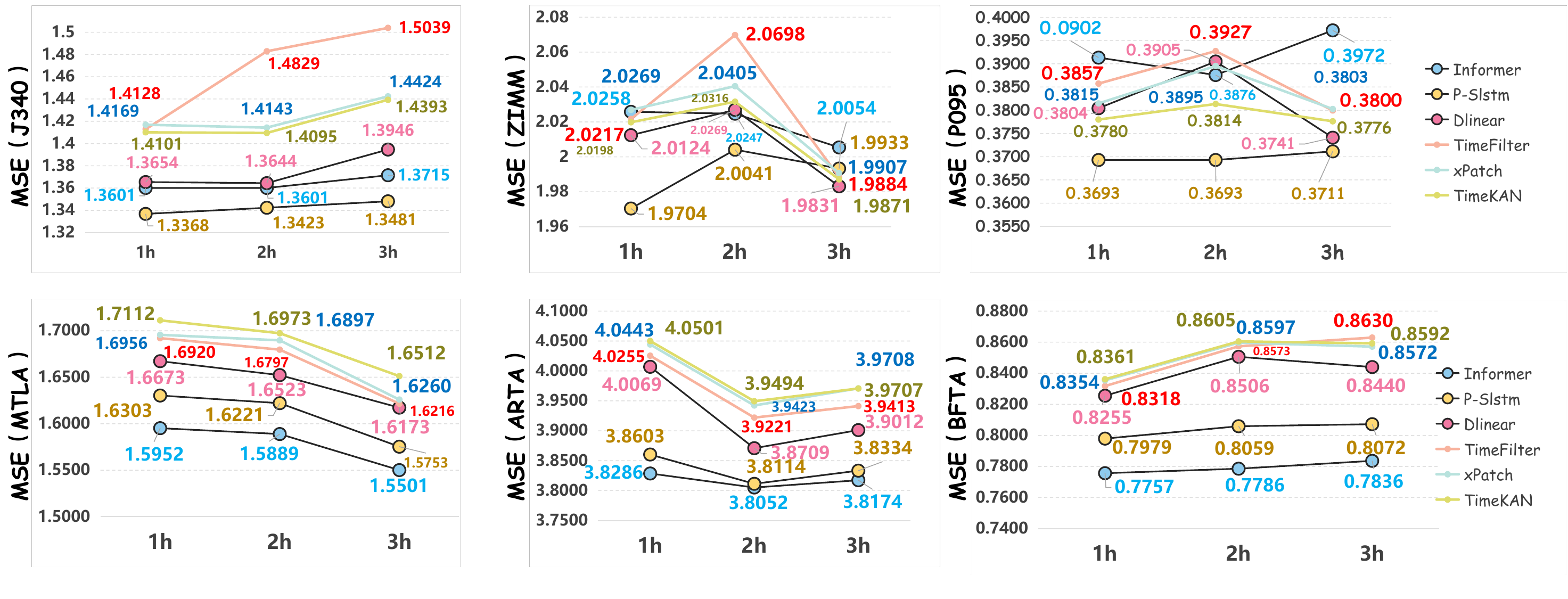}
    \caption{
Comparison of MSE results under the setting where the input resolution is fixed at 1h with a sequence length of 24. The results are shown for different output time resolutions (1h, 2h, 3h) using six representative models from different architectures.
    }
    \label{fig:multiResolution-MSE}
\end{figure}

\begin{figure}[htbp]
    \centering
    \includegraphics[width=0.9\linewidth]{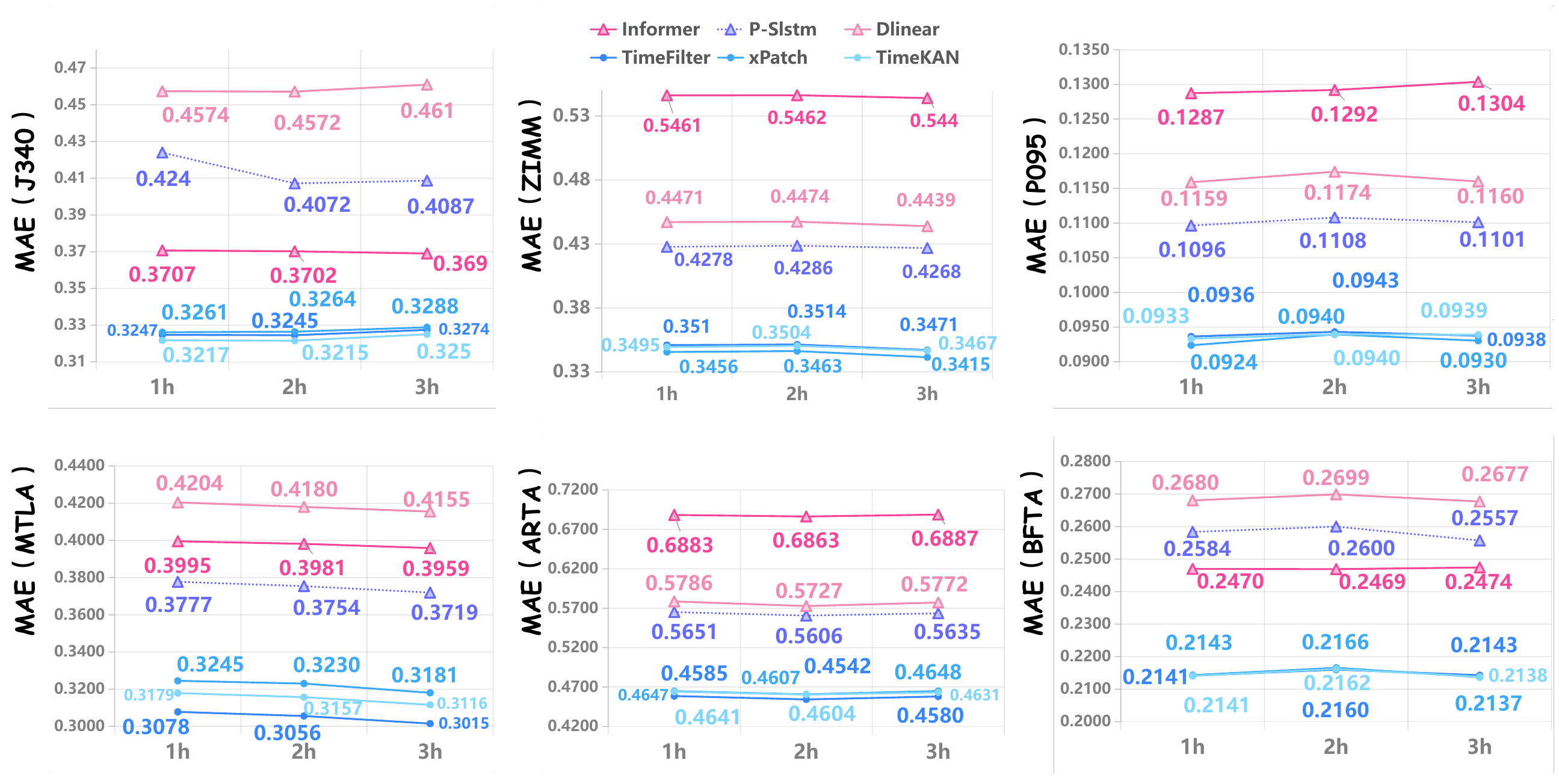}
    \caption{
Comparison of MAE results under the setting where the input resolution is fixed at 1h with a sequence length of 24. The results are shown for different output time resolutions (1h, 2h, 3h) using six representative models from different architectures.
    }
    \label{fig:multiResolution-MAE}
\end{figure}

To investigate how temporal resolution influences rainfall forecasting performance, we conduct experiments using six representative models selected from different architectural families. The input window is fixed at 24 hours, while the prediction horizon is set to 6 hours under three temporal resolutions: 1-hour, 2-hour, and 3-hour. This design allows for a fair comparison of model adaptability to varying temporal granularities while maintaining consistent historical context length. Experiments are conducted on six stations excluding FLM5, as the MSE values at FLM5 are all zero and thus not informative for analysis.

\subsubsection{ Advantages of Short Temporal Resolution}

At the 1-hour temporal resolution, all models generally exhibit superior performance at J340, P095, BFTA and ZIMM stations, reflected in lower MSE and MAE values. This indicates that finer temporal granularity allows models to respond more effectively to short-term rainfall fluctuations and rapid changes. Shorter forecasting intervals encourage the models to capture fine-grained temporal variations in precipitation, thus achieving higher predictive accuracy.

For the J340 station, where rainfall exhibits high temporal variability, the models demonstrate strong adaptability at 1-hour resolution, effectively capturing rapid fluctuations and achieving the best performance. In contrast, although rainfall patterns at the ZIMM station are relatively stable, a finer resolution still enables the models to identify subtle changes in precipitation intensity, maintaining a high level of accuracy.

\subsubsection{Adaptability at Moderate Temporal Resolution}

When the temporal resolution increases to 2 hours, model performance generally declines compared to the 1-hour setting but still retains the ability to capture long-term rainfall trends. This suggests that the 2-hour resolution strikes a balance between accuracy and stability. Certain models, such as TimeFilter and xPatch, show noticeable improvement at this resolution. Their architectures are capable of suppressing short-term noise and emphasizing more persistent rainfall patterns, thereby reducing prediction errors.

For J340, although the prediction errors increase slightly, the models demonstrate enhanced robustness by tolerating short-term fluctuations while maintaining reasonable accuracy. At ZIMM, the 2-hour setting further stabilizes the predictions by minimizing the influence of high-frequency variations, resulting in smoother and more reliable forecasts.

\subsubsection{ Advantages of Long Temporal Resolution}

At the 3-hour temporal resolution, the overall error of most models decreases, particularly at the ZIMM station. Coarser temporal aggregation enables the models to ignore short-term rainfall fluctuations and focus on broader temporal trends. Consequently, the models exhibit improved adaptability over extended forecasting horizons, producing more stable predictions.

Interestingly, for the J340 station, performance also improves at 3-hour resolution. Despite the site’s highly variable rainfall, the longer temporal window allows the models to average out transient fluctuations, resulting in greater prediction stability. For ZIMM, the 3-hour resolution achieves the best overall results, suggesting that the relatively steady rainfall regime benefits from longer-term temporal modeling, which effectively smooths short-term variability and enhances prediction precision.

\subsubsection{ Architectural Differences in Temporal Adaptability}

Distinct model architectures show varying adaptability to different temporal resolutions. Models such as Informer perform best at 1-hour resolution but degrade as the resolution increases to 2 or 3 hours, implying that Informer is more effective in capturing short-term rainfall dynamics but less capable of modeling longer temporal dependencies.

In contrast, models like TimeFilter and xPatch exhibit greater stability at coarser resolutions. Their architectural designs allow them to model long-term dependencies and attenuate the impact of transient noise, making them more suitable for long-horizon rainfall forecasting. These observations suggest that while some architectures are optimized for short-term high-frequency variability, others are inherently better at learning broader temporal structures and maintaining stability under coarser resolutions.

\subsection{RQ6: Extreme Rainfall Evaluation}

\begin{table*}[t]\setlength\tabcolsep{3pt}
\centering
\caption{Comparison of state-of-the-art methods using $\text{EERE}$  and $\text{AEERE}$. The \colorbox{red!50}{red} indicates the best-performing model, while the \colorbox{red!25}{pink} highlights the second-best. Results are obtained with an input sequence length of 24 and an output sequence length of 6.}
\resizebox{0.95\textwidth}{!}{
\begin{tabular}{cc cc cc cc cc cc cc cc cc}
\toprule
\multirow{2}{*}{Methods} & \multirow{2}{*}{\makecell{}{Publication}} & \multicolumn{2}{c}{J340} & \multicolumn{2}{c}{ZIMM} & \multicolumn{2}{c}{P095}& \multicolumn{2}{c}{MTLA}& \multicolumn{2}{c}{ARTA}&
\multicolumn{2}{c}{BFTA} & \multicolumn{2}{c}{Average} \\ 
\cmidrule(r){3-4} \cmidrule(r){5-6} \cmidrule(r){7-8} \cmidrule(r){9-10} \cmidrule(r){11-12}
\cmidrule(r){13-14} \cmidrule(r){15-16} \cmidrule(r){17-18} 
& & \makecell{$\text{EERE}$} & \makecell{$\text{AEERE}$} & \makecell{$\text{EERE}$} & \makecell{$\text{AEERE}$} & \makecell{$\text{EERE}$} & \makecell{$\text{AEERE}$} & \makecell{$\text{EERE}$} & \makecell{$\text{AEERE}$} & \makecell{$\text{EERE}$} & \makecell{$\text{AEERE}$} & \makecell{$\text{EERE}$} & \makecell{$\text{AEERE}$} & \makecell{$\text{EERE}$} & \makecell{$\text{AEERE}$} \\
\midrule
\multicolumn{10}{l}{\textbf{MLP-based}} \\ 
\midrule
DLinear & AAAI 2023 & 22.8450 & 3.6492 & 40.9955 & 4.3539 & 36.7998 & 3.7790 & 37.2634 & 4.6218 & 66.1400 & 5.3937 & 27.5221 & 4.0384 & 38.5934 & 4.3060  \\
Koopa & NIPS 2023 & 23.5072 & 3.7239 & 39.8164 & 4.2974 & 36.8885 & 3.8416 & 39.5161 & 4.8214 & 65.6357 & 5.3243 & 27.6505 & 4.0767 & 39.0677 & 4.3615 \\
FilterTS & AAAI 2025 & 23.7645 & 3.7734 & 40.7536 & 4.3703 & 36.5237 & 3.8018 & 38.8153 & 4.7354 & 65.7082 & 5.3468 & 27.9116 & 4.0942 & 38.7303 & 4.3320 \\
\midrule
\multicolumn{10}{l}{\textbf{RNN-based}} \\ 
\midrule
SegRNN & Arxiv 2023 & \cellcolor{red!25}19.6499 & \cellcolor{red!25}3.2204 & 39.8624 & 4.1736 & 45.3867 & 3.9680 & 35.8128 & 4.4679 & 62.8311 & 4.9909 & 26.2910 & 3.8960 & 37.0863 & 4.0640 \\
xLSTM & NIPS 2024 & \cellcolor{red!50}19.3018 & \cellcolor{red!50}3.1600 & 39.9339 & 4.0988 & \cellcolor{red!25}34.8533 & \cellcolor{red!25}3.6268 & \cellcolor{red!25}34.3196 & \cellcolor{red!50}4.2915 & \cellcolor{red!25}59.0556 & \cellcolor{red!25}4.7225 & \cellcolor{red!25}24.8719 & \cellcolor{red!25}3.7037 & \cellcolor{red!25}34.0477 & 3.9655  \\
P-sLSTM & AAAI 2025 & 19.9303 & 3.2200 & 40.5985 & 4.1889 & 35.6345 & 3.7077 & 36.0481 & 4.4872 & 62.2745 & 4.9337 & 26.5149 & 3.9079 & 35.9210 & 4.0313 \\
\midrule
\multicolumn{10}{l}{\textbf{TCN\&CNN-based}} \\ 
\midrule
TimesNet & ICLR 2023 & 23.9740 & 3.7370 & 39.2341 & 4.2361 & 36.9715 & 3.8178 & 37.4585 & 4.6274 & 64.8858 & 5.2981 & 27.3534 & 4.0444 & 37.8517 & 4.3444\\
TimeMixer++ & ICLR 2025 & 24.9502 & 3.8520 & 40.9137 & 4.3761 & 37.0975 & 3.8014 & 38.2259 & 4.7189 & 67.0329 & 5.4575 & 27.6310 & 4.0717 & 39.8206 & 4.4126 \\
xPatch & AAAI 2025 & 23.9499 & 3.7833 & 40.7708 & 4.3606 & 36.6973 & 3.8166 & 37.6481 & 4.6379 & 66.1276 & 5.3890 & 27.7615 & 4.0738 & 38.4236 & 4.3197 \\
\midrule
\multicolumn{10}{l}{\textbf{GNN-based}} \\ 
\midrule
MSGNet & AAAI 2024 & 23.7785 & 3.7332 & 39.2103 & 4.2475 & 35.9827 & 3.7387 & 37.4383 & 4.6237 & 65.0232 & 5.2872 & 27.7090 & 4.0821 & 38.2916 & 4.2345 \\
TimeFilter & ICML 2025 & 24.1660 & 3.8052 & 40.6665 & 4.3542 & 37.1124 & 3.8340 & 38.2567 & 4.7183 & 66.6063 & 5.4208 & 27.4514 & 4.0558 & 39.0449 & 4.3414 \\
\midrule
\multicolumn{10}{l}{\textbf{KAN-based}} \\  
\midrule
TimeKAN & ICLR 2025 & 23.9696 & 3.7912 & 40.7439 & 4.3719 & 36.5480 & 3.8163 & 38.1419 & 4.6810 & 66.2898 & 5.3931 & 27.8029 & 4.0763 & 38.7107 & 4.3500 \\
\midrule
\multicolumn{10}{l}{\textbf{Transformer-based}} \\ 
\midrule            
Informer & AAAI 2021 & 21.9020 & 3.4213 & \cellcolor{red!25}38.9007 & \cellcolor{red!25}3.9677 & 35.3396 & 3.6302 & \cellcolor{red!50}33.8138 & \cellcolor{red!25}4.3066 & \cellcolor{red!50}58.0322 & \cellcolor{red!50}4.5754 & 25.5998 & 3.8338 & 35.4618 & \cellcolor{red!50}3.8719 \\
PatchTST & ICLR 2023 & 24.3824 & 3.8322 & 40.5208 & 4.3523 & 36.5113 & 3.8016 & 38.7469 & 4.7452 & 66.9026 & 5.4353 & 27.5674 & 4.0564 & 39.6489 & 4.3627\\
iTransformer & ICLR 2024 & 24.4305 & 3.8263 & 40.6644 & 4.3322 & 36.5856 & 3.7454 & 38.4726 & 4.7286 & 67.6571 & 5.4837 & 27.5420 & 4.0520 & 39.3107 & 4.3685\\
TimeXer & NIPS 2024 & 27.2264 & 4.1162 & 43.1492 & 4.5180 & 36.7903 & 3.8786 & 41.5588 & 4.9646 & 70.0247 & 5.6137 & 29.0332 & 4.2314 & 42.0109 & 4.5796 \\
PPDformer & ICASSP 2025 & 23.7322 & 3.7836 & 40.5189 & 4.3383 & 37.0328 & 3.7920 & 38.2363 & 4.7054 & 67.7312 & 5.5073 & 27.1105 & 4.0038 & 38.6343 & 4.3117  \\
Informer(with BFPF)& Ours&23.5520&3.8909&\cellcolor{red!50}38.4074&\cellcolor{red!50}3.3550&\cellcolor{red!50}23.6849&\cellcolor{red!50}3.1865&41.9029&4.9190&58.7526&5.1080&\cellcolor{red!50}17.3577&\cellcolor{red!50}3.1483&\cellcolor{red!50}33.9437&\cellcolor{red!25}3.9346\\ 
\midrule
% Average& \textbackslash &22.8264 & 3.6869 & 40.4659 & 4.2637 & 36.8106 & 3.8150 & 38.4408 & 4.6102 & 65.5535 & 5.2189 & 27.0524 & 4.0247 & \textbackslash & \textbackslash\\
Average& \textbackslash & 23.2785 & 3.6844 & 40.3145 & 4.2385 & 36.2467 & 3.7547 & 37.8709 & 4.6557 & 64.8173 & 5.2601 & 26.7045 & 3.9693 & \textbackslash & \textbackslash\\
\bottomrule
\end{tabular}
}
\label{tab:sota_method_extream}
\end{table*}

 \begin{figure}[htbp]
    \centering
    \includegraphics[width=0.8\linewidth]{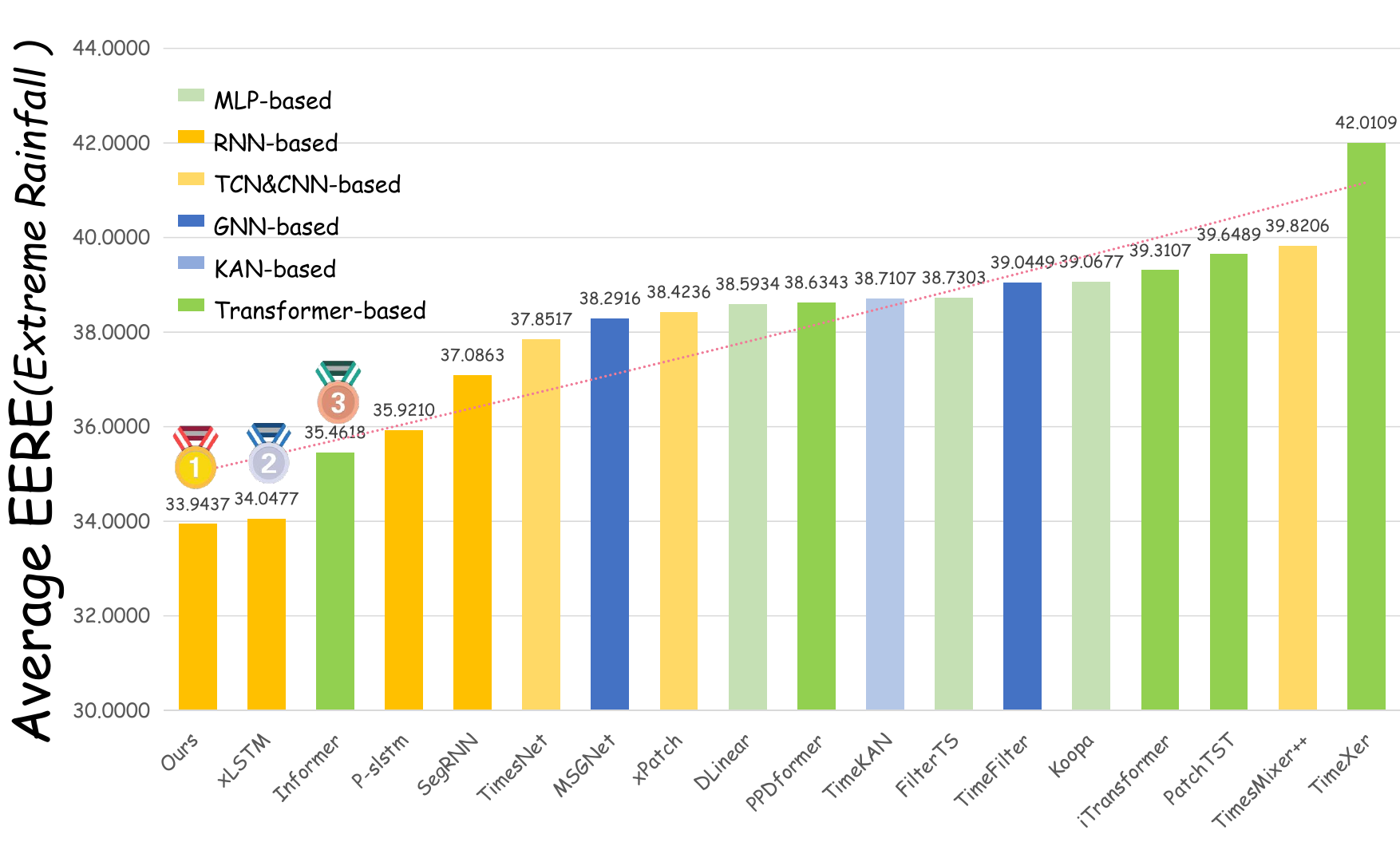}
    \caption{
       Model performance on extreme rainfall prediction. The table shows the average EERE for different models across six stations, excluding the FLM5 station due to the absence of extreme rainfall events.
    }
    \label{fig:extream_result}
\end{figure}
Comprehensive forecasting results for extream rainfall forcasting are listed in Table \ref{tab:sota_method_extream} with the best in \colorbox{red!50}{red} and the second in \colorbox{red!25}{pink}. The lower $\text{EERE}$/$\text{AEERE}$ indicates the more accurate prediction result.

Interestingly, compared with general settings (Figure \ref{fig:results}), we observe consistently higher deviation under extreme rainfall conditions, revealing the limitations of existing models and the need for advances in extreme rainfall forecasting.

 \subsubsection{RNN-based models demonstrate strong robustness under extreme conditions}
Among them, \textbf{xLSTM} achieves the lowest average \textit{EERE} (34.0477) and \textit{AEERE} (3.9655), outperforming both classical SegRNN and the probabilistic variant P-sLSTM. This suggests that the hierarchical gating and extended memory design in xLSTM effectively capture long-range dependencies and rare but impactful rainfall events.

\subsubsection{ Transformer-based architectures exhibit competitive yet inconsistent performance}  
The \textbf{Informer} model ranks second overall, with strong results on multiple stations (e.g., ZIMM and MTLA), indicating its efficiency in modeling long temporal sequences. However, other Transformer variants such as PatchTST and iTransformer show higher variance across stations, implying that self-attention alone struggles to generalize under highly sparse and extreme-valued rainfall distributions.

\subsubsection{ Models from CNN, GNN, and KAN families achieve stable but moderate performance}  
Networks like TimeFilter and TimeKAN produce consistent results across regions, but their ability to capture extreme rainfall spikes remains limited compared to recurrent structures. This indicates that local convolution and kernel-based mechanisms may fail to adequately emphasize rare temporal peaks.

\subsubsection{ Overall insights}  
Our proposed model achieves the best performance in extreme rainfall prediction, attributed to the BFPF’s enhanced capability in capturing sparse signals. Meanwhile, the relatively strong performance of xLSTM suggests that RNN-based architectures can be competitive once tailored with rainfall-specific mechanisms.

    % 画出预测值和真实值的曲线图

     % 根据前面的图分析，写上消融实验的结果
\section{Conclusion}
We introduce RainfallBench, the first benchmark tailored for GNSS-based precipitation nowcasting from the perspective of deep learning for time series forecasting, explicitly integrating PWV as a key input. Evaluating over 17 state-of-the-art time-series models, we uncover key limitations of general-purpose forecasters. To address these, we propose the Bi-Focus Precipitation Forecaster, a plug-and-play module that embeds rainfall-specific inductive biases. Results show that such domain-aware designs significantly enhance forecasting accuracy.

Future directions include exploring more effective utilization of the PWV variable, improving the model's capability in forecasting extreme events and exploring model transferability across stations.

\section*{Acknowledgments}
This work was in part supported by the Science and Technology Innovation 2030 (Grant No.2022ZD0160604), NSFC (Grant No.62176194) and Key R\&D Program of Hubei Province (Grant No.2023BAB083).

\bibliography{IEEEabrv}

\begin{thebibliography}{10}
\providecommand{\url}[1]{#1}
\csname url@samestyle\endcsname
\providecommand{\newblock}{\relax}
\providecommand{\bibinfo}[2]{#2}
\providecommand{\BIBentrySTDinterwordspacing}{\spaceskip=0pt\relax}
\providecommand{\BIBentryALTinterwordstretchfactor}{4}
\providecommand{\BIBentryALTinterwordspacing}{\spaceskip=\fontdimen2\font plus
\BIBentryALTinterwordstretchfactor\fontdimen3\font minus \fontdimen4\font\relax}
\providecommand{\BIBforeignlanguage}[2]{{%
\expandafter\ifx\csname l@#1\endcsname\relax
\typeout{** WARNING: IEEEtran.bst: No hyphenation pattern has been}%
\typeout{** loaded for the language `#1'. Using the pattern for}%
\typeout{** the default language instead.}%
\else
\language=\csname l@#1\endcsname
\fi
#2}}
\providecommand{\BIBdecl}{\relax}
\BIBdecl

\bibitem{zhang2023skilful}
Y.~Zhang, M.~Long, K.~Chen, L.~Xing, R.~Jin, M.~I. Jordan, and J.~Wang, ``Skilful nowcasting of extreme precipitation with nowcastnet,'' \emph{Nature}, vol. 619, no. 7970, pp. 526--532, 2023.

\bibitem{nearing2024global}
G.~Nearing, D.~Cohen, V.~Dube, M.~Gauch, O.~Gilon, S.~Harrigan, A.~Hassidim, D.~Klotz, F.~Kratzert, A.~Metzger \emph{et~al.}, ``Global prediction of extreme floods in ungauged watersheds,'' \emph{Nature}, vol. 627, no. 8004, pp. 559--563, 2024.

\bibitem{franch2020taasrad19}
G.~Franch, V.~Maggio, L.~Coviello, M.~Pendesini, G.~Jurman, and C.~Furlanello, ``Taasrad19, a high-resolution weather radar reflectivity dataset for precipitation nowcasting,'' \emph{Scientific Data}, vol.~7, no.~1, p. 234, 2020.

\bibitem{tang2023postrainbench}
Y.~Tang, J.~Zhou, X.~Pan, Z.~Gong, and J.~Liang, ``Postrainbench: A comprehensive benchmark and a new model for precipitation forecasting,'' \emph{arXiv preprint arXiv:2310.02676}, 2023.

\bibitem{rivero2015short}
C.~R. Rivero, H.~D. Pati{\~n}o, and J.~A. Pucheta, ``Short-term rainfall time series prediction with incomplete data,'' in \emph{2015 international joint conference on neural networks (IJCNN)}.\hskip 1em plus 0.5em minus 0.4em\relax IEEE, 2015, pp. 1--6.

\bibitem{an2025deep}
S.~An, T.-J. Oh, E.~Sohn, and D.~Kim, ``Deep learning for precipitation nowcasting: A survey from the perspective of time series forecasting,'' \emph{Expert Systems with Applications}, vol. 268, p. 126301, 2025.

\bibitem{wu2021autoformer}
H.~Wu, J.~Xu, J.~Wang, and M.~Long, ``Autoformer: Decomposition transformers with auto-correlation for long-term series forecasting,'' \emph{Advances in neural information processing systems}, vol.~34, pp. 22\,419--22\,430, 2021.

\bibitem{mouatadid2023subseasonalclimateusa}
S.~Mouatadid, P.~Orenstein, G.~Flaspohler, M.~Oprescu, J.~Cohen, F.~Wang, S.~Knight, M.~Geogdzhayeva, S.~Levang, E.~Fraenkel \emph{et~al.}, ``Subseasonalclimateusa: A dataset for subseasonal forecasting and benchmarking,'' \emph{Advances in Neural Information Processing Systems}, vol.~36, pp. 7960--7992, 2023.

\bibitem{10737421}
W.~Yin, C.~Zhou, F.~Zhou, Y.~Tian, X.~Yang, X.~Wang, R.~Tian, Y.~Xiao, W.~Zhang, J.~Kong, and Y.~Yao, ``A lightning nowcasting model using gnss pwv and multisource data,'' \emph{IEEE Transactions on Geoscience and Remote Sensing}, vol.~62, pp. 1--10, 2024.

\bibitem{liu2022non}
Y.~Liu, H.~Wu, J.~Wang, and M.~Long, ``Non-stationary transformers: Exploring the stationarity in time series forecasting,'' \emph{Advances in neural information processing systems}, vol.~35, pp. 9881--9893, 2022.

\bibitem{liu2023koopa}
Y.~Liu, C.~Li, J.~Wang, and M.~Long, ``Koopa: Learning non-stationary time series dynamics with koopman predictors,'' \emph{Advances in neural information processing systems}, vol.~36, pp. 12\,271--12\,290, 2023.

\bibitem{liutimestacker}
Q.~Liu, C.~Xu, W.~Jiang, K.~Wang, L.~Ma, and H.~Li, ``Timestacker: A novel framework with multilevel observation for capturing nonstationary patterns in time series forecasting,'' in \emph{Forty-second International Conference on Machine Learning}, 2025.

\bibitem{liutimebridge}
P.~Liu, B.~Wu, Y.~Hu, N.~Li, T.~Dai, J.~Bao, and S.-T. Xia, ``Timebridge: Non-stationarity matters for long-term time series forecasting,'' in \emph{Forty-second International Conference on Machine Learning}, 2025.

\bibitem{yao2017establishing}
Y.~Yao, L.~Shan, and Q.~Zhao, ``Establishing a method of short-term rainfall forecasting based on gnss-derived pwv and its application,'' \emph{Scientific reports}, vol.~7, no.~1, p. 12465, 2017.

\bibitem{profetto2025two}
L.~Profetto, A.~Antonini, L.~Fibbi, A.~Ortolani, and G.~M. Dimitri, ``A two-step machine learning approach integrating gnss-derived pwv for improved precipitation forecasting,'' \emph{Entropy}, vol.~27, no.~10, p. 1034, 2025.

\bibitem{10942428}
M.~Liu, W.~Zhang, Y.~Lou, X.~Dong, Z.~Zhang, and X.~Zhang, ``A deep learning-based precipitation nowcasting model fusing gnss-pwv and radar echo observations,'' \emph{IEEE Transactions on Geoscience and Remote Sensing}, vol.~63, pp. 1--9, 2025.

\bibitem{11077366}
C.~Lu, X.~Luo, Y.~Zheng, Q.~Wang, J.~Li, and Z.~Wang, ``Rsg-gan: A gan-based precipitation nowcasting model integrating radar qpe, goes-16 swd, and gnss ztds,'' \emph{IEEE Transactions on Geoscience and Remote Sensing}, vol.~63, pp. 1--17, 2025.

\bibitem{yin2024lightning}
W.~Yin, C.~Zhou, F.~Zhou, Y.~Tian, X.~Yang, X.~Wang, R.~Tian, Y.~Xiao, W.~Zhang, J.~Kong \emph{et~al.}, ``A lightning nowcasting model using gnss pwv and multi-source data,'' \emph{IEEE Transactions on Geoscience and Remote Sensing}, 2024.

\bibitem{hu2025fintsb}
Y.~Hu, Y.~Li, P.~Liu, Y.~Zhu, N.~Li, T.~Dai, S.-t. Xia, D.~Cheng, and C.~Jiang, ``Fintsb: A comprehensive and practical benchmark for financial time series forecasting,'' \emph{arXiv preprint arXiv:2502.18834}, 2025.

\bibitem{klotergens2025physiome}
C.~Kl{\"o}tergens, V.~K. Yalavarthi, R.~Scholz, M.~Stubbemann, S.~Born, and L.~Schmidt-Thieme, ``Physiome-ode: A benchmark for irregularly sampled multivariate time series forecasting based on biological odes,'' \emph{arXiv preprint arXiv:2502.07489}, 2025.

\bibitem{roque2025cherry}
L.~Roque, V.~Cerqueira, C.~Soares, and L.~Torgo, ``Cherry-picking in time series forecasting: How to select datasets to make your model shine,'' in \emph{Proceedings of the AAAI Conference on Artificial Intelligence}, vol.~39, no.~19, 2025, pp. 20\,192--20\,199.

\bibitem{aksugift}
T.~Aksu, G.~Woo, J.~Liu, X.~Liu, C.~Liu, S.~Savarese, C.~Xiong, and D.~Sahoo, ``Gift-eval: A benchmark for general time series forecasting model evaluation,'' in \emph{NeurIPS Workshop on Time Series in the Age of Large Models}, 2024.

\bibitem{qiu2024tfb}
X.~Qiu, J.~Hu, L.~Zhou, X.~Wu, J.~Du, B.~Zhang, C.~Guo, A.~Zhou, C.~S. Jensen, Z.~Sheng \emph{et~al.}, ``Tfb: Towards comprehensive and fair benchmarking of time series forecasting methods,'' \emph{Proceedings of the VLDB Endowment}, vol.~17, no.~9, pp. 2363--2377, 2024.

\bibitem{bkaczek2023tspp}
J.~B{\k{a}}czek, D.~Zhylko, G.~Titericz, S.~Darabi, J.-F. Puget, I.~Putterman, D.~Majchrowski, A.~Gupta, K.~Kranen, and P.~Morkisz, ``Tspp: A unified benchmarking tool for time-series forecasting,'' \emph{arXiv preprint arXiv:2312.17100}, 2023.

\bibitem{liu2023largest}
X.~Liu, Y.~Xia, Y.~Liang, J.~Hu, Y.~Wang, L.~Bai, C.~Huang, Z.~Liu, B.~Hooi, and R.~Zimmermann, ``Largest: A benchmark dataset for large-scale traffic forecasting,'' \emph{Advances in Neural Information Processing Systems}, vol.~36, pp. 75\,354--75\,371, 2023.

\bibitem{de2021rainbench}
C.~S. de~Witt, C.~Tong, V.~Zantedeschi, D.~De~Martini, A.~Kalaitzis, M.~Chantry, D.~Watson-Parris, and P.~Bilinski, ``Rainbench: Towards data-driven global precipitation forecasting from satellite imagery,'' in \emph{Proceedings of the AAAI Conference on Artificial Intelligence}, vol.~35, no.~17, 2021, pp. 14\,902--14\,910.

\bibitem{shi2017deep}
X.~Shi, Z.~Gao, L.~Lausen, H.~Wang, D.-Y. Yeung, W.-k. Wong, and W.-c. Woo, ``Deep learning for precipitation nowcasting: A benchmark and a new model,'' \emph{Advances in neural information processing systems}, vol.~30, 2017.

\bibitem{deng2021lenghu}
L.~Deng, F.~Yang, X.~Chen, F.~He, Q.~Liu, B.~Zhang, C.~Zhang, K.~Wang, N.~Liu, A.~Ren \emph{et~al.}, ``Lenghu on the tibetan plateau as an astronomical observing site,'' \emph{Nature}, vol. 596, no. 7872, pp. 353--356, 2021.

\bibitem{papee1961chemical}
H.~M. Pap{\'e}e and A.~Montefinale, ``Chemical composition of precipitable water vapour over the united states,'' \emph{Nature}, vol. 191, no. 4784, pp. 136--138, 1961.

\end{thebibliography}
\bibliographystyle{IEEEtran}

\end{document}